\documentclass[aps,prx,reprint,superscriptaddress,longbibliography,floatfix]{revtex4-2}
\usepackage{amsmath,amssymb}
\usepackage{graphicx}
\usepackage{xcolor}
\usepackage[colorlinks=true,allcolors=blue]{hyperref}

\colorlet{RED}{red} 

\begin{document}

\title{A preference for local interaction between layers emerges during training}
\author{Jiheon Lee}
\email{anffl0101@kias.re.kr}
\affiliation{Quantum Universe Center, Korea Institute for Advanced Study, Seoul 02455, South Korea}
\date{\today}

\begin{abstract}
A deep neural network is a stack of layers, and all layers are trained to lower one shared error. A change in one layer can therefore reach every other layer. 
Because the error is shared by all layers, one would expect the other layers to respond alike, whether they are next to the changed layer or far from it. We find that early in training the response is weak and its order by distance is not stable. As training proceeds, the response grows, and neighboring layers come to respond to changes in each other more strongly than layers further apart, in almost every network. This locality forms during training, although training links every layer to every other. 
We show this by pushing one layer of a copy of the network at many stages of training and measuring how much of the push reaches each of the other layers. The experiments use 512 five-layer networks trained on modular addition with each of two optimizers, AdamW and stochastic gradient Langevin dynamics.
\end{abstract}

\maketitle

\section{Introduction}
\label{sec:intro}

\begin{figure*}[!t]
    \centering
    \includegraphics[width=0.8\textwidth]{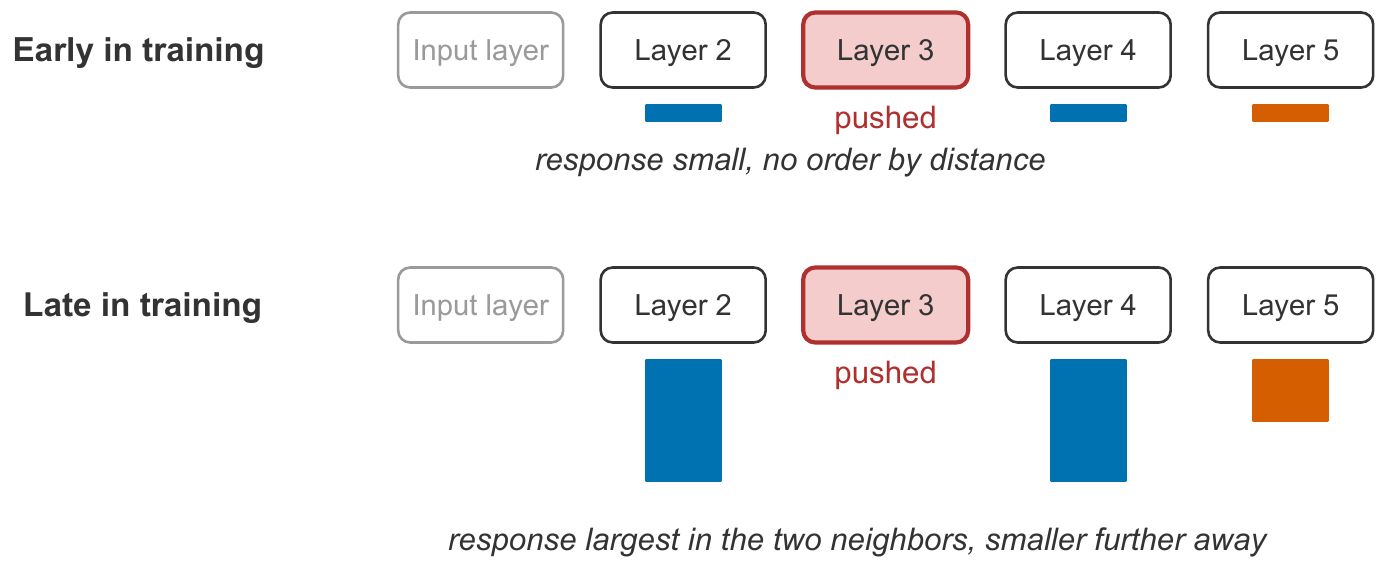}
    \caption{Sketch of the main result. One layer of a copy of the network is pushed (red), and the bars show how strongly the other layers respond (blue for the neighbors, orange for the layer two apart). Early in training the response is small and is not ordered by the distance from the pushed layer. Late in training the neighbors of the pushed layer respond most, and layers further away respond less. The first layer, labeled input layer, is left out of the analysis.}
    \label{fig:sketch}
\end{figure*}

A deep network computes its answer by passing numbers from one layer to the next, but it is trained in a different way. Every layer is changed to lower one shared error (training loss), so the update of each layer is connected to all the other layers, i.e., a change in one layer can reach any other layer. Much is known about what single layers learn~\cite{zeiler2014visualizing,yosinski2014transferable,raghu2017svcca}, and about how the loss of the whole network drops during training~\cite{saxe2014exact,jacot2018neural}, and how the loss of language models drops as they grow~\cite{kaplan2020scaling}. However, little is known about how the layers influence each other while the network learns. Does a change in one layer reach all other layers alike, or prefer its neighbors? And does this change during training?

Neither answer is obvious, because a network does two different things. When it computes an answer, each layer feeds only the next one, so the first layer reaches the third only through the second. Here neighbors are special by construction. When it learns, every layer is corrected from the same error at the output, so a change of one layer changes the very next update of any other layer directly, without any change of the layers in between. So in learning, unlike in computing, neighbors are not special by construction. Whether they become special during training is the question we study.

We measure this influence directly. We copy a network during training, push one layer of the copy for a short time, and then train the copy and the untouched network side by side under the same conditions, so that any difference between them is caused by the push. We then measure how much of the push reaches each of the other layers. Such comparisons of a perturbed copy with the untouched network have been made on neural networks before~\cite{altintas2025butterfly,lin2026mapping}, but not to follow a change from one layer to the others.

To see whether the influence changes during training, we need networks that are small enough to be easily trained, and whose weights keep changing over a long training run. Small networks trained on modular arithmetic, which are known to grok~\cite{power2022grokking,liu2022omnigrok}, give this. They fit their training examples early, and their internal structure keeps forming for a long time afterward~\cite{nanda2023progress}. Grokking thus gives us a long stretch of training in which the network keeps changing, so that we can investigate how it evolves over time.

Studies of grokking have looked inside the network~\cite{nanda2023progress,yunis2024spectral,clauw2024information}, but not at how the layers affect each other's learning. Earlier studies of the influence between layers measure something else. Most ask how the layers work together to compute the answer. During training, Refs.~\cite{mattsson2020examining,moldovan2023information} measured how much information passes from one layer to the next. In a transformer that groks, replacing the output of one layer at one input position by its value for another input can change the network's answer, an effect that grows for some layers during grokking~\cite{wang2024grokked}. In language models, earlier studies skipped one block, i.e., a group of layers, or made the outputs of two parts larger or smaller together, at saved points of training, to see how much the answers change~\cite{tao2026persistent,ji2026bridging}. None of these measurements follows how an effect propagates over training steps. Each is made on the network at one stage of training, and repeating it at later stages gives a series of snapshots of how the network answers. So they see the influence in computing, not the influence in learning. Fewer studies ask, as we do, how the layers affect each other's learning, but they mostly use a change that lasts the whole run. Keeping the first or the last layer at its starting weights makes the other layers, above all its neighbors, do less to lower the loss~\cite{lan2019lca}. Neurons on neighboring layers appear to come to depend on each other during training~\cite{yosinski2014transferable}. Training together with the upper layers, those closer to the output, can correct what the lower layers have learned~\cite{allenzhu2023backward}. A change that lasts from the first step to the last gives one result for the whole run, so it cannot show when the influence appears or how it changes as the network learns.

In our setup, we use one of the tasks of the first study of grokking, adding two numbers and keeping the remainder after division by a prime number~\cite{power2022grokking}. We train 512 networks with five layers with each of two optimizers. The first is AdamW~\cite{loshchilov2019decoupled}, which is widely used in practice. The second is stochastic gradient Langevin dynamics (SGLD)~\cite{welling2011bayesian}, which is used in theoretical studies of training~\cite{raginsky2017nonconvex}.

Our main finding is that the layers develop a local interaction during training. (Schematics in Fig.~\ref{fig:sketch}.) Late in training, a push on one layer reaches its neighbors most, layers two apart less, and layers three apart least, on average over the pairs at each distance. This holds in almost every AdamW network and in more than nine of ten SGLD networks. We call an interaction with this order local in distance. This order of the response is not built into the network itself. Early in training, a push reaches the other layers only weakly and without a stable order by distance. This matters for how deep networks can be understood. A system of many parts in which each part interacts mainly with its neighbors is far simpler to describe than one in which every part acts on every other, and it is the kind of system that theoretical physics has the most tools for. Our results suggest that a network moves toward this simpler form during training.

Earlier work has described how the layers of a network relate to each other. In transformers and residual networks, the blocks come to align as a network trains~\cite{aubry2025coupling,li2023residual}, in one large language model most strongly between neighboring blocks~\cite{aubry2025coupling}. This alignment concerns how the blocks pass the signal forward, not how one block affects the learning of another. Other quantities between layers favor neighboring layers by construction~\cite{lee2025bulk} or already at the start in networks without skip connections~\cite{bolshim2026interlayer,dejong2025emergent}. The influence that we measure, of the strongest patterns of one layer on the learning of another layer, is instead not local early in training and becomes local as training proceeds. To our knowledge, no earlier study has found an influence of one layer on the learning of another that becomes ordered by the distance between layers during training.

We also find that the line-up of neighboring layers, which is known from earlier work~\cite{saxe2014exact,ji2018aligns,yunis2024spectral,yunis2026thesis,jeong2026geometric}, also in a related form in transformers~\cite{bray2026common}, is tied to the local interaction. Two neighboring layers line up when the strongest output patterns of one layer become the strongest input patterns of the next. Among networks at the same training step, neighbors that line up more respond more strongly to each other, in the SGLD networks at every step and in the AdamW networks while their line-up forms.

For SGLD we derive how a line-up makes two layers respond to each other in the first step after a push. When two neighboring layers line up, a change of either one changes the network's answers in a similar way, so training responds to a push on one of them by changing the other. Using earlier results on how the layers of a network stay balanced during training~\cite{arora2018optimization,du2018algorithmic,le2022training,rawal2026saddle,nystrom2026commutator}, we then derive which neighbors line up. Weight decay lines them up where the neurons between them switch on and off together. Equations for the balance of layers are known for networks with the rectified linear unit (ReLU)~\cite{le2022training} and with weight decay~\cite{nystrom2026commutator}. What is new is the use of the rule to say which neighbor pairs line up and which do not, and the test of this on three neighbor pairs of the SGLD networks.

The paper is organized as follows. We first describe the task, the network, and the two training methods in Sec.~\ref{sec:setup}. The push experiment is described in Sec.~\ref{sec:method}. Our results, including why neighboring layers respond to each other most, are in Sec.~\ref{sec:results}. We summarize and give an outlook in Sec.~\ref{sec:conclusion}. Details of the methods and the full calculation are given in the appendices.

\section{Setup}
\label{sec:setup}

The networks learn $(a+b)\bmod p$ with $p=113$. The input is the pair $(a,b)$, written as two vectors of length 113, each with a single 1 at the position of the number and 0 elsewhere, placed one after the other. The output is 113 scores, one for each possible answer, and the predicted answer is the one with the highest score. Of the $113^2=12\,769$ possible questions, 40\% are the training examples. They are chosen at random once and are the same for all networks. The other 60\% are the test examples. The loss $L$ is the cross-entropy of the scores,
\begin{equation}
L\equiv-\frac1N\sum_x\log P(x),\qquad P(x)\equiv\frac{e^{y_c(x)}}{\sum_k e^{y_k(x)}},
\label{eq:loss}
\end{equation}
where the score $y_k(x)$ is the output of the last layer for answer $k$ ($k=0,\dots,112$), and $c=(a+b)\bmod p$ is the correct answer.
Here $x=(a,b)$ is one training example, and the sum runs over all $N=5108$ of them. The number $P(x)$ is then the probability that the network gives to the correct answer. The loss is never negative, and it approaches 0 as $P(x)$ approaches 1 for every training example.

The network is a fully connected network with five layers of weights: $W_1$ ($256\times226$), $W_2$--$W_4$ ($256\times256$), and $W_5$ ($113\times256$), each with a bias $b_i$. The activation function is the ReLU, $\varphi(z)\equiv\max(0,z)$, applied to each neuron after each of the first four layers.
For one example $x$, the network computes, one layer after the other,
\begin{equation}
\begin{aligned}
z_i&=W_ih_{i-1}+b_i,\quad h_i=\varphi(z_i)\qquad(i=1,\dots,4),\\
y&=W_5h_4+b_5,
\end{aligned}
\label{eq:network}
\end{equation}
starting from $h_0=x$, the input of length 226. Here $z_i$ holds the inputs of the 256 neurons after $W_i$ and before the ReLU, $h_i$ their outputs, and $y=(y_0,\dots,y_{112})$ the scores.
Before training, each weight matrix is set to a random orthogonal matrix (for the non-square $W_1$ and $W_5$, a random matrix with orthonormal columns or rows), and $W_1$ to $W_4$, whose outputs go through a ReLU, are then multiplied by $\sqrt2$. The biases start at zero. 

Training changes the weights and biases in small steps so that the loss $L$ decreases. The rule that sets each step is called the optimizer. We use two optimizers, AdamW~\cite{loshchilov2019decoupled} and SGLD~\cite{welling2011bayesian} with a fixed step size and a temperature~\cite{raginsky2017nonconvex}, and train 512 networks with each. In both, every step uses the gradient of $L$ over all training examples, not over a small random batch of them. Both methods also use weight decay, which pulls every weight and bias toward zero. Each step shrinks them by the factor $1-\eta\lambda$, where $\eta$ is the step size and $\lambda$ the weight decay. For SGLD, this shrinking is the same as adding the term $\frac{\lambda}{2}\sum_\theta\theta^2$ to the loss, where $\theta$ runs over every entry of the weight matrices $W_i$ and the bias vectors $b_i$. AdamW applies the shrinking directly instead of through the loss~\cite{loshchilov2019decoupled}, because otherwise it would divide this term by the typical gradient size as well. In Ref.~\cite{loshchilov2019decoupled} the step size does not multiply the weight decay, so the weight decay there corresponds to $\eta\lambda$ here.

AdamW is a widely used optimizer. It sets a separate step size for each weight from the size of its recent gradients, so it trains many kinds of networks well without much tuning, and it is the standard choice for large language models. It keeps for each weight a running mean $m$ of its gradient and a running mean $v$ of its squared gradient. At step $t$, for every weight and bias $\theta$ with gradient $g=\partial L/\partial\theta$,
\begin{equation}
\begin{aligned}
m&\to\beta_1m+(1-\beta_1)\,g,\qquad v\to\beta_2v+(1-\beta_2)\,g^2,\\
\theta&\to(1-\eta\lambda)\,\theta-\eta\,\frac{\hat m}{\sqrt{\hat v}+\epsilon},
\end{aligned}
\label{eq:adamw}
\end{equation}
where $m$ and $v$ start at zero, $\hat m\equiv m/(1-\beta_1^t)$ and $\hat v\equiv v/(1-\beta_2^t)$ correct for this start, and $\epsilon$ is a small number that keeps the divisor away from zero. So each weight's step is its mean gradient divided by its typical gradient size. This holds as long as that size is large compared with $\epsilon$. We use $\eta=10^{-3}$, $\lambda=1$, $\epsilon=10^{-8}$, and $\beta_1=0.9$ and $\beta_2=0.999$ for the running means. The same $\eta$ and $\lambda$ were used in an earlier study of grokking on the same arithmetic task~\cite{varma2023explaining}. Each of the 512 AdamW networks starts from its own random orthogonal weights.

SGLD adds Gaussian noise to every gradient step. It is used in theoretical studies of training, because it is a discretized form of the Langevin equation, whose long-run behavior is known for suitable losses~\cite{raginsky2017nonconvex}. Each step updates every weight and bias $\theta$ by
\begin{equation}
\theta\to(1-\eta\lambda)\,\theta-\eta\,\frac{\partial L}{\partial\theta}+\sqrt{2\eta T}\,\xi,
\label{eq:sgld}
\end{equation}
where $\xi$ is a Gaussian number of mean 0 and variance 1, drawn for each weight and bias at each step. We use $\eta=0.1$, $\lambda=10^{-3}$, and the temperature $T=3\times10^{-5}$. Equation~\eqref{eq:sgld} is the Langevin equation $d\theta=-(\partial U/\partial\theta)\,dt+\sqrt{2T}\,dB$, taken in time steps $dt=\eta$, with the energy $U\equiv L+\frac\lambda2\sum_\theta\theta^2$ and Brownian motion $B$. For small steps the weights would in the long run be spread as $e^{-U/T}$, the Boltzmann distribution at temperature $T$ with the loss plus weight decay as the energy. We do not assume that our networks reach it. All 512 SGLD networks start from the same weights and differ only in their noise.

All sets were trained on NVIDIA A10 GPUs. The weights are kept in 32-bit floating point. For AdamW the probabilities of the answers (the softmax of the scores, as in Eq.~\eqref{eq:loss}) and the loss are computed in 64-bit floating point, because in 32 bits rounding in the softmax can disturb training~\cite{prieto2025grokking,liu2026grokking}. SGLD uses 32 bits in the whole process. The AdamW networks were trained for 150\,000 steps and the SGLD networks for 200\,000 steps, long after the training examples were fitted in both. The training and test accuracy and the stable rank of each layer, defined below, were logged at every step up to step 300 and every 100 steps after that for AdamW, and every 20 steps for SGLD. All random numbers come from recorded seeds, so every run can be repeated exactly on the same type of GPU with the same software and the same number of CPU threads.

We describe each layer by its stable rank. By the singular value decomposition (SVD), every weight matrix can be written as a sum of patterns, $W_i=\sum_k\sigma_ku_kv_k^{\mathrm T}$. Here the input patterns $v_k$ and the output patterns $u_k$ are orthonormal vectors with $W_iv_k=\sigma_ku_k$, and the strengths $\sigma_1\ge\sigma_2\ge\dots\ge0$ are the singular values of $W_i$. The stable rank is defined by
\begin{equation}
r_i\equiv \frac{\|W_i\|_F^2}{\sigma_1^2}=\frac{\sum_k\sigma_k^2}{\sigma_1^2},
\label{eq:srank}
\end{equation}
where $\|W_i\|_F^2$, the squared Frobenius norm, is the sum of the squares of all entries of $W_i$ and equals the sum of the squared strengths. So $r_i$ is the total squared strength of all patterns divided by that of the strongest one. It is 1 when all the weight lies in one pattern, and it drops when a few patterns grow stronger than the rest. We chose it for four reasons. First, it does not change when the neurons of a hidden layer are merely permuted, or when one layer is multiplied by a number and the next divided by it. Second, it has a simple meaning. It is roughly the number of patterns among which the weight is spread. Third, it is a smooth function of the weights wherever the strongest pattern is strictly stronger than the second ($\sigma_1>\sigma_2$), so a layer can be pushed along its gradient. Fourth, it is one number per layer and cheap to compute, so it can be recorded for every network during training.

Unless stated otherwise, the analyses below use all 512 networks of each optimizer, and they compare the networks at the same training step. The caption of each figure and table gives the number of networks that enter it. Uncertainties are 95\% bootstrap intervals. The networks are drawn again with replacement many times, the quantity is computed for each draw, and the interval holds the central 95\% of the resulting values.

\section{Measuring the influence between layers}
\label{sec:method}

We measure the influence of one layer on the others by changing its weights. At a chosen training step we copy the network, push one layer of the copy, and then train the copy and the untouched network, under identical conditions, so that every later difference between the two runs is caused by the push. We first define the push and its realization for each optimizer. We then define the quantities read from the two runs, the one-way response and the mutual response.

A push on layer $i$ moves its weights along the gradient of its stable rank $r_i$,
\begin{equation}
W_i\to W_i+\kappa\,\frac{\partial r_i}{\partial W_i},\qquad \frac{\partial r_i}{\partial W_i}=\frac{2}{\sigma_1^2}\left(W_i-r_i\,\sigma_1u_1v_1^{\mathrm T}\right),
\label{eq:push}
\end{equation}
where $\kappa>0$ sets the size of the push, and $\sigma_1$, $u_1$, and $v_1$ belong to the strongest pattern of $W_i$. Among all infinitesimal moves of a given size, this one raises $r_i$ the most. The two optimizers need different realizations of it. (The derivation of the gradient in Eq.~\eqref{eq:push} and the form and size of the push for each optimizer are in Appendix~\ref{app:push}.)

For AdamW, copies are made every 1000 steps from step 4000 to step 30\,000, and every 5000 steps from there to step 145\,000. They start at step 4000, after layers 2--5 have lost most of the directions of their random start. The mean stable rank of these layers falls from 256 (113 for layer 5) to about 6--7 by this step and stays in this range afterward. At each of these steps every layer is pushed in a copy of its own. The push is applied over the first 32 updates of the copy by
\begin{equation}
W_i\to W_i+\frac{\alpha}{32}\,\|W_i\|_F\,\frac{\partial r_i/\partial W_i}{\|\partial r_i/\partial W_i\|_F}.
\label{eq:pushadamw}
\end{equation}
This is Eq.~\eqref{eq:push} with $\kappa=\alpha\,\|W_i\|_F/(32\,\|\partial r_i/\partial W_i\|_F)$, where the strength $\alpha$ of the push is its size as a fraction of $\|W_i\|_F$. We use $\alpha=0.1$ unless stated otherwise. Every push moves its layer by the same fraction of its size $\|W_i\|_F$, so the responses to different pushes can be compared. The push is added after the usual AdamW update, and the running means $m$ and $v$ do not enter.

For SGLD, copies are made every 5000 steps from step 10\,000 to step 150\,000. For the first 32 updates of the copy, the loss $L$ is replaced by $L-h\,r_i$, where $h$ is the strength of the push. The pushed layer is then updated by
\begin{equation}
W_i\to(1-\eta\lambda)\,W_i-\eta\,\frac{\partial L}{\partial W_i}+\sqrt{2\eta T}\,\xi+\eta h\,\frac{\partial r_i}{\partial W_i}.
\label{eq:pushsgld}
\end{equation}
This is Eq.~\eqref{eq:push} with $\kappa=\eta h$. The copy draws the same noise $\xi$ as the untouched network. We use $h=0.1$ unless stated otherwise. This push moves the pushed layer by an estimated 6--17\% of its size $\|W_i\|_F$. The size $\alpha=0.1$ of the AdamW push was chosen inside this range, so that the pushes of the two methods have a similar size.

The response of layer $j$ to a push on layer $i$ is $\Delta r_j^{(i)}(t,\tau)\equiv \tilde r_j(t+\tau)-r_j(t+\tau)$, the difference in stable rank between the pushed copy ($\tilde r_j$) and the untouched network ($r_j$), at the delay $\tau$, that is, after $\tau$ updates of a copy that is made at step $t$ and has a push on layer $i$. We define the one-way response of layer $j$ to this push as its response divided by the change of the pushed layer $i$,
\begin{equation}
\chi_{j\leftarrow i}(t,\tau,\tau')\equiv \frac{\langle\Delta r_j^{(i)}(t,\tau)\rangle}{\langle\Delta r_i^{(i)}(t,\tau')\rangle},
\label{eq:pairresponse}
\end{equation}
where $\langle\cdot\rangle$ is the mean over all pushes on layer $i$ made inside one bin of $t$, and $\tau'$ is the delay at which the pushed layer is read. The bins of $t$ are 5000 training steps wide. The pushed layer is read either at the same delay, $\tau'=\tau$, or at the end of the push, $\tau'=32$. We call the push in the first of the 32 updates the first push, so $\tau=2$ is one step and $\tau=512$ is 511 steps after it.

For a pair of layers we define the mutual response as the geometric mean of the two one-way responses,
\begin{equation}
\chi_{ij}(t,\tau,\tau')\equiv \sqrt{\left|\chi_{j\leftarrow i}(t,\tau,\tau')\,\chi_{i\leftarrow j}(t,\tau,\tau')\right|}.
\label{eq:coupling}
\end{equation}
The geometric mean makes $\chi_{ij}$ the same in both directions.

The training accuracy of the SGLD networks falls from 1 to about 0.2 roughly once every 1700--3000 steps per network and recovers within a few hundred steps. We call such a fall a collapse. A gradient step of size $\eta$ is stable only while the largest curvature of the loss is below $2/\eta$, and full-batch gradient descent is known to drive the curvature up to this limit~\cite{cohen2021edge}, with a spike of the loss when the curvature is above it~\cite{cohen2021edge,lewkowycz2020catapult}. Here the largest curvature is the largest eigenvalue of the Hessian of the loss, the matrix of its second derivatives with respect to the weights and biases. In the SGLD networks the largest curvature rises above $2/\eta$ shortly before most of the collapses that we examined, so the collapses are consistent with such spikes (Appendix~\ref{app:crash}). Without the noise they are about 10 times rarer. A push can also cause a collapse in the pushed copy. The untouched AdamW networks never collapse, and a push causes a collapse only rarely. 

A collapse in either network would generate a large error in our measurements. Therefore, a pushed copy and its untouched network are discarded from the first logged step at which the training accuracy of either is below 0.5. We call this rule the crash filter. For SGLD it leaves enough pushed copies only up to about 1000--2000 steps after the push, so we read the response at $\tau=512$. Appendix~\ref{app:crash} also gives the fraction of pushed copies that the filter keeps.

\section{Results}
\label{sec:results}

We first show how the mutual response depends on the distance between layers early in training, for pushes made up to step 15\,000, and late in training, for pushes made from step 60\,000 on. We read the response 511 steps and one step after the first push. We then ask why the response is local. In this section, the pairs among layers 2--5 are grouped by distance into neighbors (2--3, 3--4, and 4--5), layers two apart (2--4 and 3--5), and layers three apart (2--5). Layer 1 is not included in the analysis, since it is the only layer whose input is not computed by the network. Layers 2--5, on the other hand, read the output of another layer, so both their input and their output vary during training, and they have this structure in common.

\subsection{The interaction becomes local during training}
\label{sec:local}

\begin{figure*}[!t]
    \centering
    \includegraphics[width=\textwidth]{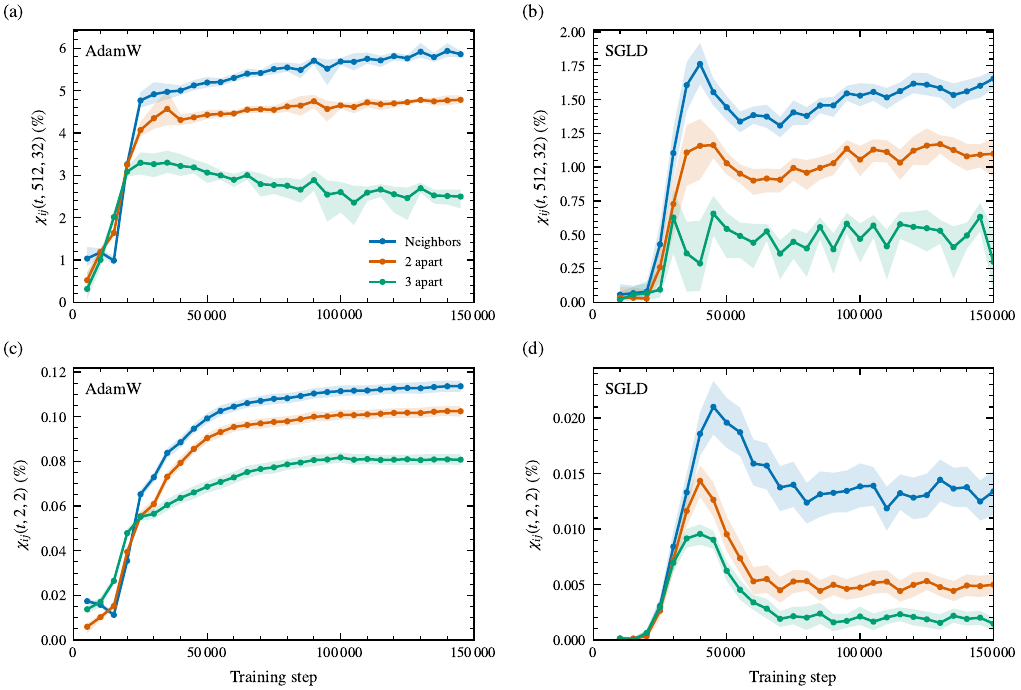}
    \caption{Mutual response by distance against the training step of the push, for AdamW (left) and SGLD (right). (a), (b) Mutual response $\chi_{ij}(t,512,32)$, 511 steps after the first push. (c), (d) Mutual response $\chi_{ij}(t,2,2)$, one step after the first push. Pairs are grouped by distance, for pushes at $\alpha=0.1$ for AdamW and at $h=0.1$ for SGLD. Each point includes only the networks with a pushed copy that passes the crash filter. For AdamW these are 511 or 512 networks at every point. For SGLD they are 232--388 of the 512 networks at $\tau=512$, and 479--504 at $\tau=2$. Bands show the 95\% bootstrap interval.}
    \label{fig:adamw}
\end{figure*}

Figure~\ref{fig:adamw} shows the mutual response between layers against the training step of the push. Each point is a mean over the pushed copies of all 512 networks that pass the crash filter. For AdamW the filter removes almost none of the pushed copies, and for SGLD it removes 52\% at $\tau=512$ and 3\% at $\tau=2$. The caption of Fig.~\ref{fig:adamw} gives the numbers for each point.

Early in training the response is small and has no stable order by distance. Later, a push on one layer reaches its neighbors most, layers two apart less, and layers three apart least, with both methods. This is an order of the mean mutual response over the pairs at each distance, and we call it the full order [Fig.~\ref{fig:adamw}(a), (b)]. In the AdamW networks it holds from step 20\,000 on, and in the SGLD networks from step 25\,000 on. We also repeated the experiment with different push strengths, which give the same rise of the mutual response and the same order by distance after it (Appendix~\ref{app:strength}).

The mutual response at $\tau=512$ can include effects that have passed through other layers. We therefore also plot $\chi_{ij}(t,2,2)$, which shows which layers change which others directly one step after the first push [Fig.~\ref{fig:adamw}(c), (d)]. This response is much smaller, because it is not built up over many updates, and its order by distance is weaker. However, it still shows the same trend. In the AdamW networks the order holds from step 25\,000 on, and in the SGLD networks from step 30\,000 on. In AdamW the order appears in the same steps in which $\sqrt{\hat v}$ falls from far above $\epsilon$ to far below it (Appendix~\ref{app:alike}).

After $\sqrt{\hat v}$ has fallen below $\epsilon$, the same change of the gradient gives a much larger step, so this change of AdamW could itself be the source of the order. However, it does not produce the order. We computed the one-step mutual response for an infinitely small push from the saved weights of the 512 AdamW networks, once for the AdamW step and once for a plain gradient step. At step 10\,000 the full order holds in 24\% of the networks for the AdamW step and in 28\% for the plain gradient step, and neighbors respond more than layers three apart in only 49\% and 46\% of the networks, as for a random order. From step 60\,000 on it holds in 91--94\% and 81--87\%. So the order does not come from the divisor $\sqrt{\hat v}+\epsilon$ of AdamW.

Table~\ref{tab:fullorder} checks the full order in each network separately. For each network it compares the pushes made early in training with those made late in training. The table uses every network that still has, after the crash filter, a push on every layer both early and late. We chose the two limits so that the steps in which the response changes most, between steps 15\,000 and 60\,000, fall outside both windows (Fig.~\ref{fig:adamw}). At $\tau=512$ the full order holds late in training in 99.6\% of the AdamW and 92.1\% of the SGLD networks, against 30.3\% and 26.1\% early in training. One step after the first push it holds in 97.5\% and 82.4\%, against 23.2\% and 39.3\%. If the two limits are moved to step 10\,000 and to step 40\,000 or 80\,000, each of these fractions changes by at most 7 percentage points. 

The full order is a strict requirement. Even where it fails, the preference for near layers remains. Late in training, at $\tau=512$, neighbors respond to each other more than layers three apart in 99.8\% of the AdamW networks and in 99.4\% of the SGLD networks of Table~\ref{tab:fullorder}. 

 The pairs two and three apart, 2--4, 3--5, and 2--5, all contain layer 2 or layer 5, so this order could also come from the overall strength of these two layers rather than from the distance. With the strength of each layer removed, neighbors still respond more than layers two apart late in training in 99.4\% of the AdamW and in 72.7\% of the SGLD networks at $\tau=512$, against 61.7\% and 52.1\% early (Appendix~\ref{app:strengthfree}). So the interaction between layers becomes local during training, with neighbors responding most in almost every network.

In the AdamW networks, the local interaction does not require a network to generalize. In 14 of the 512 AdamW networks the test accuracy never exceeds 0.03, although all of them fit every training example. Late in training all 14 are in the full order at $\tau=512$ (Appendix~\ref{app:nogen}). These networks never learn to answer the test examples, and the local interaction still emerges in them during training. The local interaction is therefore not specific to grokking, which suggests that it is a more general feature of training.

\begin{table}[!t]
    \centering
    \caption{Fraction of networks that are in the full order early and late in training. In each network the mutual response is computed from the pushes made up to step 15\,000 (Early) and from step 60\,000 on (Late). The table uses the networks that pass the filter on every layer, both early and late. These are all 512 AdamW networks, and 165 SGLD networks at $\tau=512$ and all 512 at $\tau=2$. The rows give the fraction of networks in the full order of $\chi_{ij}(t,512,32)$ ($\tau=512$) and of $\chi_{ij}(t,2,2)$ ($\tau=2$). The 95\% bootstrap interval is given in parentheses below each fraction.}
    \label{tab:fullorder}
    \begin{tabular}{lcccc}
        \hline\hline
         & \multicolumn{2}{c}{AdamW} & \multicolumn{2}{c}{SGLD} \\
         & Early (\%) & Late (\%) & Early (\%) & Late (\%) \\
        \hline
        $\tau=512$ & 30.3 & 99.6 & 26.1 & 92.1 \\
         & (26.4--34.2) & (99.0--100.0) & (19.4--32.7) & (87.9--95.8) \\[3pt]
        $\tau=2$ & 23.2 & 97.5 & 39.3 & 82.4 \\
         & (19.7--27.0) & (96.1--98.8) & (35.2--43.6) & (79.1--85.7) \\
        \hline\hline
    \end{tabular}
\end{table}

\subsection{Neighbors that line up respond to each other}
\label{sec:why}

We showed that neighbors respond to each other most strongly late in training. We now ask why it happens. 

We first ask by which route a push reaches another layer. In both methods the update of layer $j$ depends on the other layers only through the gradient
\begin{equation}
\frac{\partial L}{\partial W_j}=\frac1N\sum_x J_j(x)^{\mathrm T}q(x),
\end{equation}
where $J_j(x)\equiv \partial y(x)/\partial W_j$, and $q(x)\equiv -\partial\log P(x)/\partial y(x)$ is the error on example $x$, the 113 probabilities that the network gives to the answers minus 1 at the correct answer. A small push $\Delta W_i$ on layer $i$ changes the gradient by~\cite{martens2020new}
\begin{equation}
\Delta\left(\frac{\partial L}{\partial W_j}\right)=\frac1N\sum_x\Bigl[J_j^{\mathrm T}\frac{\partial q}{\partial y}\,J_i\Delta W_i+(\Delta J_j)^{\mathrm T}q\Bigr].
\label{eq:routes}
\end{equation}
This is how the push on layer $i$ changes the gradient for the next step of layer $j$. The first term measures the overlap of how the error $q$ is changed by the push on layer $i$ with how layer $j$ moves the scores. It is the block of the generalized Gauss--Newton matrix that links layers $i$ and $j$, applied to the push $\Delta W_i$, and for the cross-entropy of the scores the generalized Gauss--Newton matrix equals the Fisher matrix~\cite{martens2020new}. The response one step after the push thus reads this block along the two push directions. The second term is the overlap of the error $q$ with the change that the push on layer $i$ makes directly in $J_j$. The two terms point to different causes of the local interaction. If the first term is the larger one, neighbors respond most because their changes move the scores alike. If the second term is the larger one, the cause lies in how the signal passes between the layers.

We compute the two terms for both methods. Table~\ref{tab:routes} shows the Frobenius norm of the second term of Eq.~\eqref{eq:routes} divided by that of the first term. The push $\Delta W_i$ is along the gradient of the stable rank $r_i$, as in the previous experiments. In AdamW the step of each weight is normalized by its own $\sqrt{\hat v}+\epsilon$, which is set by its past gradients, and we include this factor of each weight in both terms. At every step in the table this ratio is 1.6--2.0\% for AdamW and 5--16\% for SGLD. The first term is thus the dominant source of the response.

\begin{table}[!t]
    \centering
    \caption{Frobenius norm of the second term of Eq.~\eqref{eq:routes} divided by that of the first term, at three training steps and from step 60\,000 on. The squared norms are summed over the pairs. The SGLD column has no entry at step 10\,000, because the weights of the SGLD networks were saved every 20\,000 steps. All 512 AdamW networks enter in every row. For SGLD, 507, 491, and 512 networks survive the crash filter in the three rows, respectively. The 95\% bootstrap interval is given in parentheses.}
    \label{tab:routes}
    \begin{tabular}{lcc}
        \hline\hline
         & AdamW (\%) & SGLD (\%) \\
        \hline
        Step 10\,000 & 1.6 & -- \\
         & (1.5--1.6) & \\[3pt]
        Step 20\,000 & 2.0 & 16.3 \\
         & (1.9--2.0) & (12.7--19.5) \\[3pt]
        Step 40\,000 & 1.8 & 4.9 \\
         & (1.7--1.8) & (4.4--5.4) \\[3pt]
        From step 60\,000 & 1.7 & 7.0 \\
         & (1.6--1.7) & (5.9--8.4) \\
        \hline\hline
    \end{tabular}
\end{table}

One step after a push, the other layers therefore respond mainly to the change that the push makes in the error. This response is largest in a layer that moves the scores in a similar way to the pushed layer. The mutual response can then be local if neighboring layers come to move the scores more alike than layers further apart do. The ReLU itself links neighboring layers, since the incoming weights and the bias of a hidden neuron can be doubled and its outgoing weights halved without changing the scores. This link is in the network from the start of training, whereas the response becomes local only during training. So this link alone does not make the response local.

Neighboring layers can move the scores alike when they are lined up. Suppose that the strongest output pattern of $W_i$ and the strongest input pattern of $W_{i+1}$ are the same and that the ReLU passes this pattern unchanged. If the bias of layer $i$ along the shared pattern is neglected, the output of layer $i+1$ for training example $x$ along this pattern is the product $\sigma_1(W_{i+1})\,\sigma_1(W_i)\,a(x)$, where $a(x)$ is the component of the input of layer $i$ along the strongest input pattern of $W_i$. A push on either layer along its strongest pattern changes one factor of this product, by the same fraction for every example, so the two changes of the scores are proportional (Appendix~\ref{app:alike}),
\begin{equation}
\delta y_{i+1}(x)=c_i\,\delta y_i(x),
\label{eq:alike}
\end{equation}
where $\delta y_i(x)$ is the change of the scores of training example $x$ that a push on layer $i$ makes, per push size, and $c_i$ is one positive number, the same for all 113 scores and every training example. The two layers thus can move the scores alike.

The derivation makes three simplifications, and all three hold well in our networks late in training (Appendix~\ref{app:alike}). First, the line-up is a property of the strongest patterns of two layers, but the push of Eq.~\eqref{eq:push} moves every pattern of the pushed layer. This is a good approximation, since the strongest pattern holds about 85\% of the squared size of the push in the AdamW networks and 96--98\% in the SGLD networks, both from step 60\,000 on. Second, the bias of layer $i$ is neglected. This also works, since the mean of the squared bias along the strongest pattern is at most 0.02\% of $\sigma_1^2$ in the AdamW networks and at most 0.7\% in the SGLD networks. Third, a minimal gradient step is assumed, in which every weight moves by the same multiple of its gradient. For AdamW, from step 30\,000 on, the divisor $\sqrt{\hat v}+\epsilon$ is nearly the same for most weights, so the change that a push makes in the next step of most weights is nearly the same multiple of the change of its gradient. At earlier steps the derivation does not strictly apply.

The same argument covers layers that are further apart. The signal of layer $i$ reaches layer $j$ only through the layers between them. The two layers therefore move the scores alike only as far as every pair of neighbors on the way does, and what is lost at one pair is not restored at the next. The cosine between the changes of the scores of layers $i$ and $j$, taken with the inner product of the first term of Eq.~\eqref{eq:routes}, then drops with the distance between the layers, so the response is largest between neighbors. In its simplest form, this argument predicts that the cosine of two layers is the product of the cosines of the neighbor pairs between them. The AdamW networks follow this prediction. For layers 2 and 5 the cosine is 0.77, and the product is 0.76 (means over the 512 networks from step 60\,000 on). In the SGLD networks the cosine of layers further apart is larger than the product but smaller than that of every neighbor pair between them. The cosine thus falls with the distance in both methods, as fast as predicted in the AdamW networks and more slowly in the SGLD networks. The product rule holds if two layers are linked only through the layers between them. 

We test this with the partial correlation of two layers. From the change of the scores made by each of the two layers, we remove the part that can be written as a combination of the changes made by the other two layers, and take the cosine of what is left. This cosine is zero if the two layers are alike only through the other layers. It is obtained for all pairs at once from the inverse of the matrix of cosines of layers 2--5. In the AdamW networks, from step 60\,000 on, it is 0.18--0.94 for neighbors. Of the pairs that are not neighbors, the pairs 3--5 and 2--5 have a partial correlation of zero within the bootstrap interval, and the pair 2--4 has $-0.12$. At step 20\,000 the pairs that are not neighbors still have up to 0.42. So during training the AdamW layers come to be linked directly mainly to their neighbors, although the cosines fall only slowly with the distance. In the inverse of the matrix of cosines, this means that the entries away from the diagonal are largest for neighbors and small for the other pairs, a structure similar to that found for the inverse of the Fisher matrix~\cite{martens2015kfac}. In the SGLD networks the partial correlation is 0.06--0.19 for neighbors and 0.01--0.04 for the pairs that are not neighbors. Appendix~\ref{app:alike} gives the measured cosines by distance for both optimizers.

Therefore, we measured how well neighboring layers line up. The line-up of layers $i$ and $i+1$ is defined as $|u_1(W_i)\cdot v_1(W_{i+1})|$, where $u_1(W_i)$ is the strongest output pattern of $W_i$ and $v_1(W_{i+1})$ is the strongest input pattern of $W_{i+1}$. It is 1 when the two patterns are the same and about 0.05 for two unrelated patterns, that is, two random patterns of 256 entries.

\begin{figure*}[!t]
    \centering
    \includegraphics[width=\textwidth]{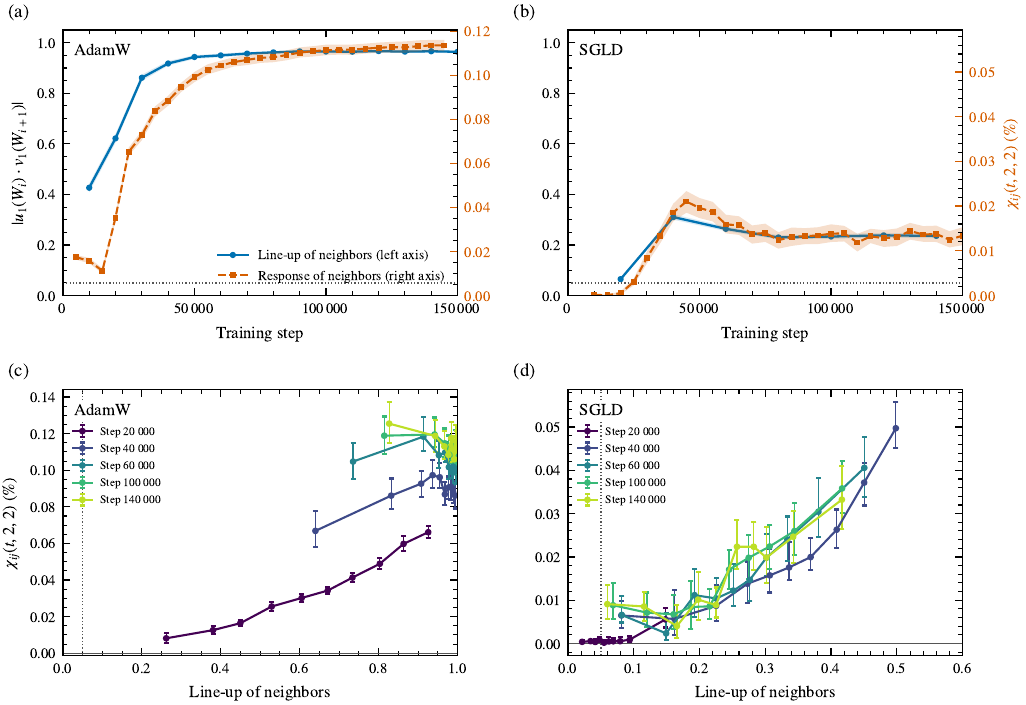}
    \caption{Mean line-up of neighboring layers and their mutual response over the networks, for AdamW (left) and SGLD (right). (a), (b) Line-up of the strongest patterns, $|u_1(W_i)\cdot v_1(W_{i+1})|$, averaged over the three neighbor pairs (solid line, left axis), and mutual response of neighbors one step after the first push, $\chi_{ij}(t,2,2)$ (dashed line, right axis), against the training step. The scale of the right axis is chosen so that the two curves agree from step 60\,000 on, to compare their shapes. (c), (d) Mutual response of neighbors one step after the first push, $\chi_{ij}(t,2,2)$, against the line-up of neighbors. At each training step, the networks are sorted by their line-up, averaged over the three neighbor pairs, and cut into ten groups of equal size. Each curve shows the ten groups sorted by line-up at each step, with 49--52 networks in a group. The dotted lines mark unrelated patterns (0.050). The line-up in (a), (b) and the curves in (c), (d) use the saved states with training accuracy at least 0.5, as in Table~\ref{tab:routes}, with 512 AdamW and 491--507 SGLD networks at each step. The response in (a), (b) is the neighbor curve of Fig.~\ref{fig:adamw}(c), (d), with the same networks. Bands around the response in (a), (b) and error bars in (c), (d) show the 95\% bootstrap interval. For the line-up in (a), (b) this interval is at most $\pm 0.02$.}
    \label{fig:lineup}
\end{figure*}

Figure~\ref{fig:lineup}(a), (b) shows the line-up averaged over the three neighbor pairs, together with the mutual response of neighbors one step after the first push. In the AdamW networks the line-up rises steeply up to step 30\,000 and stays close to 1 from step 50\,000 on. The response also rises in a similar way. In the SGLD networks the response rises with the line-up as well. Both have their maximum near step 40\,000 and slightly drop and maintain nonzero values afterward.

Figure~\ref{fig:lineup}(c), (d) shows how the mutual response of neighbors depends on their line-up for several training steps. For both optimizers, the networks whose neighbors are lined up more also have a larger mutual response, in the AdamW networks up to step 40\,000. In the SGLD networks the trend is not clear while the line-up is below 0.2, but it becomes clear for a larger line-up. Late in training, the line-up is close to 1 in most AdamW networks, and their response no longer rises with it. The SGLD networks fluctuate more because of the noise, so some of them still have a small line-up there. In the SGLD networks a clear relation between the line-up and the mutual response of neighbors nevertheless remains in such a regime. 

Across the networks at the same training step, the Pearson correlation between the line-up and the response of neighbors is 0.80 for AdamW at step 20\,000, 0.14 at step 40\,000, and between $-0.16$ and $-0.08$ later, and 0.36--0.62 for SGLD at the steps used in the panels. Appendix~\ref{app:pairs} shows the correlation for each neighbor pair by itself. There, the response of layers 2 and 3 of the SGLD networks, the pair that does not line up, hardly rises with the line-up. This shows that the line-up can be weaker for some pairs of layers, and the response of such a pair then does not follow it. Appendix~\ref{app:offshare} discusses why this pair does not line up. The average over the pairs shows the overall trend, whereas the levels of single pairs should not be compared with each other (Appendix~\ref{app:pairs}).

\subsection{Why neighboring layers line up}
\label{sec:balance}

To see why neighboring layers line up, consider hidden layer $i$ with $n$ neurons followed by the ReLU. From the weights we build two $n\times n$ matrices, $A_i\equiv W_{i+1}^{\mathrm T} W_{i+1}$ and $B_i\equiv \overline W_i\overline W_i^{\mathrm T}$, where $\overline W_i\equiv [\,W_i\mid b_i\,]$ is $W_i$ with the bias $b_i$ as one more column. Written with the patterns of the SVD, $A_i=\sum_k\sigma_k^2v_kv_k^{\mathrm T}$ holds the input patterns $v_k$ of $W_{i+1}$, and $B_i=\sum_k\sigma_k^2u_ku_k^{\mathrm T}$ holds the output patterns $u_k$ of $\overline W_i$, each with its squared strength. So if $A_i=B_i$, the two layers are lined up. One SGLD step changes the difference $D_i\equiv A_i-B_i$ from its value $D_i(t)$ to 
\begin{equation}
\begin{aligned}
D_i(t+1) ={}& (1-\eta\lambda)^2 D_i(t)-\eta(1-\eta\lambda)\,(K_i+K_i^{\mathrm T})\\
&+\eta^2 Q_i+R_i,
\end{aligned}
\label{eq:balance}
\end{equation}
with
\begin{equation}
\begin{aligned}
(K_i)_{kl}&=\frac1N\sum_x e_{i,k}(x)\,z_{i,l}(x)\,\bigl(m_{i,l}(x)-m_{i,k}(x)\bigr),\\
Q_i&\equiv g_{i+1}^{\mathrm T}g_{i+1}-\overline g_i\overline g_i^{\mathrm T},\\
R_i&\equiv \sqrt{2\eta T}\,(\Gamma_i+\Gamma_i^{\mathrm T})+2\eta T\,(\xi_{i+1}^{\mathrm T}\xi_{i+1}-\overline\xi_i\overline\xi_i^{\mathrm T}),\\
\Gamma_i&\equiv M_{i+1}^{\mathrm T}\xi_{i+1}-\overline\xi_i\overline M_i^{\mathrm T},
\end{aligned}
\label{eq:balance-terms}
\end{equation}
where $k$ and $l$ are the indices of the neurons, and the switch $m_{i,l}(x)$ is 1 if neuron $l$ is on, $z_{i,l}(x)>0$, and 0 if it is off. The entry $e_{i,k}(x)\equiv \partial\ell(x)/\partial h_{i,k}(x)$ is the derivative of the loss, with $\ell(x)\equiv-\log P(x)$. Also, $g_{i+1}\equiv \partial L/\partial W_{i+1}$ and $\overline g_i\equiv \partial L/\partial\overline W_i$. The matrices $M_{i+1}\equiv (1-\eta\lambda)W_{i+1}-\eta g_{i+1}$ and $\overline M_i\equiv (1-\eta\lambda)\overline W_i-\eta\overline g_i$ are the weights after one SGLD step, from $t$ to $t+1$, without its noise term. The matrices $\xi_{i+1}$ and $\overline\xi_i$ hold the Gaussian numbers $\xi$ of Eq.~\eqref{eq:sgld} for $W_{i+1}$ and $\overline W_i$. Appendix~\ref{app:lineup} derives Eq.~\eqref{eq:balance} for any activation function.

Equation~\eqref{eq:balance} says two things. First, weight decay shrinks the difference by the factor $(1-\eta\lambda)^2$ at every step. Second, of the three terms that add to the difference, only $K_i$ can keep the patterns of the two layers apart. The term $\eta^2Q_i$ is of second order in the step size $\eta$, and the noise $R_i$ adds on average a multiple of the identity matrix, which adds the same amount to every squared strength and does not change the patterns. Where $K_i$ is small, $A_i$ therefore approaches $B_i$ up to a multiple of the identity matrix, that is, the input patterns of $W_{i+1}$ become the output patterns of $\overline W_i$. The entry $(K_i)_{kl}$ is zero whenever neurons $k$ and $l$ are on together or off together on every training example, as was shown in a narrower setting~\cite{le2022training}, and $K_i=0$ in a network without the ReLU. For SGLD, two neighboring layers therefore line up where the neurons of their strongest patterns switch on and off together. For networks without the ReLU and without biases this is the known balance of layers, i.e., $A_i=B_i$, which is kept once it holds when training uses very small steps, both without weight decay~\cite{arora2018optimization,du2018algorithmic} and with it~\cite{arora2018optimization}. Earlier forms of this rule for ReLU networks are given in Ref.~\cite{le2022training}. The equation for $D_i$ without the noise was derived earlier for networks without biases, in the limit of small steps~\cite{nystrom2026commutator}.

At every step weight decay removes the fraction $2\eta\lambda$ of $D_i$, and the ReLU term adds $-\eta(K_i+K_i^{\mathrm T})$, both to first order in $\eta$. The two balance at $D_i=-(K_i+K_i^{\mathrm T})/(2\lambda)$. On the strongest output patterns of $W_i$, the matrix $A_i$ has the squared strength of $B_i$ plus this difference. When the difference is small compared with the squared strength, a pair therefore lines up, because layer $i+1$ then reacts most strongly to the output patterns of layer $i$. The term $K_i$ is large when the strongest output patterns of $W_i$ include many neurons that are off on some examples. The matrix $B_i$ counts everything that layer $i$ sends to its neurons, whether they are on or off. Layer $i+1$ receives only the outputs of the neurons that are on, because the ReLU sets the others to zero, so its input patterns cannot match these output patterns. This is the case for layers 2 and 3 of the SGLD networks, the pair that does not line up (measured in Appendix~\ref{app:offshare}).

The line-up of the SGLD networks is only partial, and it appears only after step 20\,000. Both are expected from the noise that SGLD adds to the weights. A learned pattern is among the strongest patterns of a layer only once it is stronger than the noise, and it is then still mixed with noise (Appendix~\ref{app:noise}). 

For AdamW the weight-decay part holds as it stands, but the gradient part is harder to analyze. AdamW divides the running mean $\hat m$ of the gradient of each weight by $\sqrt{\hat v}+\epsilon$, which depends on the past gradients of that weight. We therefore derive the rule only for SGLD, where every step is a gradient step, and test in the AdamW networks whether its consequences hold beyond that case. At late steps the derivation still holds approximately for AdamW. From step 30\,000 on, the divisor $\sqrt{\hat v}+\epsilon$ is nearly equal to $\epsilon$ for most weights, so the gradient part of each step is nearly $\eta/\epsilon$ times $\hat m$. Equation~\eqref{eq:balance} then holds approximately with $\hat m$ in place of the gradient, with $\eta/\epsilon$ in place of $\eta$ in the gradient terms, and without the noise term. The line-up, however, also forms at earlier steps, where the derivation does not apply. This suggests that the conclusion of the derivation holds under weaker conditions as well, but a proof of this is beyond the scope of this work.

Taken together, these results show how the local interaction appears during training in four steps. First, weight decay shrinks every weight at every step, and only the directions that the gradient keeps feeding survive. Each layer is left with a few strong patterns~\cite{yunis2024spectral}, as was also found for the map from the neurons to the scores in a one-layer transformer that groks~\cite{nanda2023progress}. Second, in the same way, weight decay removes the patterns that only one of two neighboring layers has and keeps those that they share [Eq.~\eqref{eq:balance}], so the strong patterns of neighbors line up. Third, two neighbors that are lined up change the scores alike [Eq.~\eqref{eq:alike}]. Fourth, layers that change the scores alike respond to each other through the first term of Eq.~\eqref{eq:routes}, so the response is strongest between neighbors. In our five-layer networks, however, the layers do not contribute equally to the averages by distance (e.g., layer 5 is in three of the six pairs, and layers 2 and 5 form the only pair three apart). Deeper networks have several pairs at each distance and can show how the response depends on the distance alone. We leave this for future work.

\section{Summary and outlook}
\label{sec:conclusion}

The main result is that the response between layers is not local by construction and becomes local during training. We have pushed each layer of five-layer ReLU networks trained with AdamW and SGLD optimizers and measured how much of each push reaches the other layers. Early in training, the response to a push is small and has no stable order by distance. Late in training, a push reaches the neighbors most, layers two apart less, and layers three apart least, on average over the pairs at each distance. This order holds in almost every AdamW network and in more than nine of ten SGLD networks, although the two methods differ in the step rule, the noise, and the initialization. The two methods play complementary roles. AdamW, which is widely used in practice, tests the result under the conditions in which networks are usually trained. SGLD, a discretized form of the Langevin equation, gives a simple setting in which every step is a gradient step with noise, so that the line-up of neighboring layers can be derived (Sec.~\ref{sec:balance}). 
In both methods the local order comes with a line-up of the strongest patterns of neighboring layers, large in the AdamW networks and partial in the SGLD networks, where layers 2 and 3 do not line up. At the same training step, neighbors that line up more respond more strongly to each other on average over the three neighbor pairs, in the SGLD networks at every step and in the AdamW networks while their line-up forms. We show that this is because neighboring layers that are lined up move the scores alike.

In other words, the layers acquire a short-range interaction that was not designed. The architecture of the network is the same throughout training, so this interaction is a property of the trained weights, not of how the network is designed. 
The local interaction is not equally strong between all neighbors. Some pairs respond to each other more weakly than others. The order by distance nevertheless emerges at the level of the whole network.

We believe these results give an outlook beyond the examples studied here. In fundamental physics, a short-range interaction is usually assumed from the beginning. Here a system in which every layer depends on all the others through the loss acquires such an interaction through its own dynamics. Locality across depth is thus an outcome of training. The number of layers between two layers becomes a distance for the dynamics only through training. If the response falls quickly with this distance in deeper networks, the description of how a deep network learns can be greatly simplified. It could then be approximated by keeping only the local interactions between neighboring layers, whose number grows in proportion to the depth, not to its square.

\begin{acknowledgments}
We used Claude Opus~5.5 and Fable~5.2 for assistance with code development and manuscript revision, and all AI-assisted outputs were verified. This work is supported by a KIAS Individual Grant (Grant No. QP110101) via the Quantum Universe Center at Korea Institute for Advanced Study. We are grateful to the Center for Advanced Computation at Korea Institute for Advanced Study for providing the computing resources.

\end{acknowledgments}

\section*{Data availability}
The data and scripts necessary to reproduce the numerical results and figures are available in Ref.~\cite{lee2026repository}.

\appendix

\section{Validation of the methods}
\label{app:details}

\subsection{Derivation of the push formula}
\label{app:push}

Here we derive the gradient of the stable rank $r_i=\|W_i\|_F^2/\sigma_1^2$ in Eq.~\eqref{eq:push} and the form and size of the push for each optimizer. The largest singular value is $\sigma_1=u_1^{\mathrm T}W_iv_1$, where $u_1$ and $v_1$ are the strongest output and input patterns of $W_i$. They are also the unit vectors that make $u^{\mathrm T}W_iv$ largest. The change of $u_1$ and $v_1$ does not change $\sigma_1$ to first order, so $\Delta\sigma_1=u_1^{\mathrm T}\,\Delta W_i\,v_1$ and $\partial\sigma_1/\partial W_i=u_1v_1^{\mathrm T}$. The quotient rule then gives
\begin{equation}
\frac{\partial r_i}{\partial W_i}=\frac{2W_i}{\sigma_1^2}-\frac{2\|W_i\|_F^2}{\sigma_1^3}\,u_1v_1^{\mathrm T}=\frac{2}{\sigma_1^2}\left(W_i-r_i\,\sigma_1u_1v_1^{\mathrm T}\right),
\label{eq:srgrad}
\end{equation}
the matrix in Eq.~\eqref{eq:push}. With $\sigma_1^2=\|W_i\|_F^2/r_i$ this gives 
\begin{equation}
\left\|\frac{\partial r_i}{\partial W_i}\right\|_F=\frac{2\,r_i\sqrt{r_i-1}}{\|W_i\|_F}.
\label{eq:srgradnorm}
\end{equation}
So the size of the gradient depends on the stable rank and the size of the layer, and it differs from layer to layer and changes during training.

For SGLD, the push is made by adding the term $-h\,r_i$ to the loss. In Langevin dynamics the loss plus weight decay is the energy, and adding $-h$ times a quantity to the energy is the usual way to push that quantity with a strength $h$ and measure the response to the push~\cite{Cugliandolo_1993}. The stable rank $r_i$ depends only on $W_i$, so the new term changes only the update of $W_i$, by $\eta h\,\partial r_i/\partial W_i$. This gives Eq.~\eqref{eq:pushsgld}.

The strength $h$ is fixed, but the size of the gradient is not, so the size of the SGLD push relative to the weights changes with the layer and the time. Over the 32 updates the push moves $W_i$ by about $32\,\eta h\,\|\partial r_i/\partial W_i\|_F$. By Eq.~\eqref{eq:srgradnorm}, this gives the push size $64\,\eta h\,r_i\sqrt{r_i-1}/\|W_i\|_F$. With the means of $r_i$ and $\|W_i\|_F$ over the networks in our experiments, and neglecting the training during the 32 updates, the ratio of the push size to $\|W_i\|_F$ is 6--17\% for $h=0.1$.

If the same term $-h\,r_i$ were added to the loss in AdamW training, the push on each weight would be divided by that weight's own number $\sqrt{\hat v}+\epsilon$ in Eq.~\eqref{eq:adamw}, which is set by its past gradients. The move would then not be along $\partial r_i/\partial W_i$, and its size would depend on the history of each weight. We therefore leave the AdamW update unchanged and add the move of Eq.~\eqref{eq:pushadamw} to the weights after it, so that every layer can be moved by the same fraction $\alpha$ of its size $\|W_i\|_F$ at every time.

\subsection{Crash filter in SGLD}
\label{app:crash}

Here we give the background on the crash filter through a curvature measurement at the collapses of the SGLD networks and also report the fraction of pushed copies that the filter keeps. In our experiment, there are about 46\,000 collapses in the 512 runs, about one per 1700--3000 steps per network, at random times. 

The collapses are consistent with the loss spikes of gradient descent at its stability limit. Full-batch gradient descent with a fixed step $\eta$ is stable only while the largest curvature of the loss stays below $2/\eta$, which for us is 20, and full-batch gradient descent is known to drive the curvature up to this limit~\cite{cohen2021edge}. 

\begin{figure*}[!t]
    \centering
    \includegraphics[width=\textwidth]{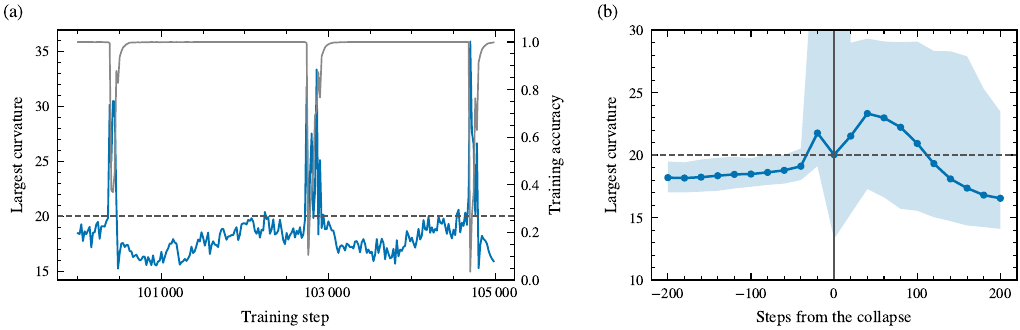}
    \caption{Collapses and the largest curvature of the loss in 508 of the 512 SGLD networks, along their training from step 100\,000 to step 105\,000. The dashed line is the stability limit $2/\eta=20$ of the fixed step. (a) One example of a collapsed network. The largest curvature of the training loss (blue, left axis) and the training accuracy (gray, right axis). (b) The largest curvature against the steps from a collapse, where the dots are the medians over all 946 collapses. The shaded band spans the 10th to 90th percentiles over the collapses. The vertical line marks the collapse. Near it the band extends beyond the top of the axis, to 41 at 20 steps before the collapse and to 121 at the collapse.}
    \label{fig:sharpness}
\end{figure*}

We measured the largest curvature of the training loss every 20 steps in 508 of the 512 SGLD networks, those whose median test accuracy over the last 10\,000 steps of training is at least 0.8, along their training from step 100\,000 to step 105\,000, by power iteration on the Hessian (Fig.~\ref{fig:sharpness}). At the points at which the training accuracy is at least 0.99, the curvature is mostly below the limit. Its median is 16.4, and it exceeds 20 at only 2.5\% of these points. There are 946 collapses in these steps, in 491 of the networks, where a fall counts as a new collapse only if the training accuracy stayed at least 0.5 during the 200 steps before it. In 733 of the collapses (77\%), the curvature rises above 20 during the 60 steps before the collapse. The median of the curvature over all collapses rises from 18.2 at 200 steps before the collapse to 21.8 at the last logged point before it, stays at 20--23 for the next 100 steps, and is 16.5 at 200 steps after the collapse. 

A push can cause a collapse in the copy that the untouched network does not have, and the difference this makes would be counted as a response to the push. So a pushed copy and its untouched network are kept in the experiments in the main text only if the training accuracy of neither has fallen below 0.5 at any logged step up to the delay $\tau$. At $h=0.1$ the fraction of pushed copies kept 511 steps after the first push, counted over the pushes on all five layers including layer 1, is 39\% for pushes made up to step 15\,000, 39\% for pushes made between steps 15\,000 and 60\,000, and 46\% for pushes made from step 60\,000 on. At 5000 steps after the push begins it is 0.1\%, 0.0\%, and 1.9\%. So almost no pushed copy is kept for as long as 5000 steps. The 512 untouched AdamW networks never fall below a training accuracy of 0.5 once they have reached 0.99. A push causes such a fall in only 206 of the 128\,000 pushed copies (0.16\%), and the filter removes these data.

\subsection{Robustness of the local interaction to the push strength}
\label{app:strength}

\begin{figure*}[!t]
    \centering
    \includegraphics[width=\textwidth]{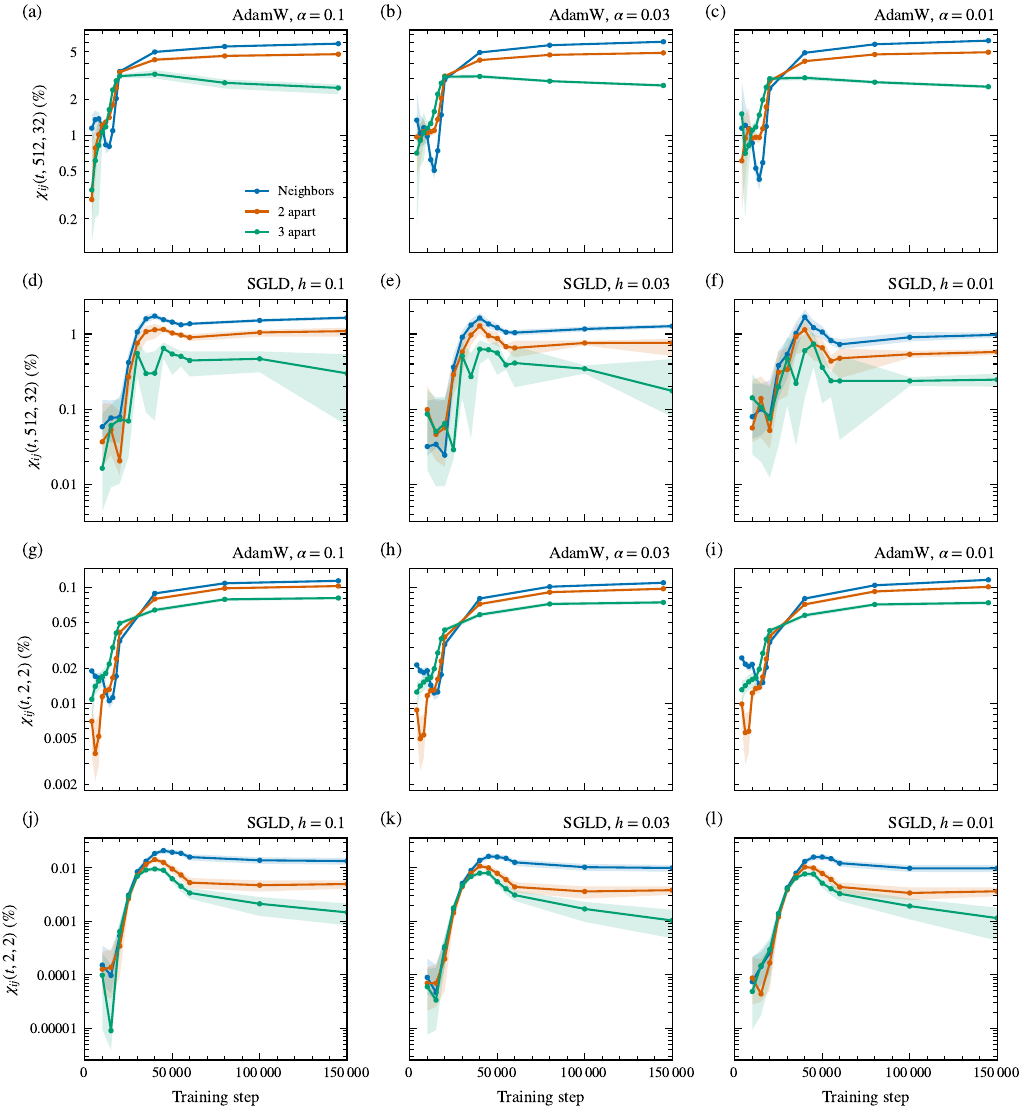}
    \caption{Mutual response by distance for three push strengths, for AdamW at $\alpha=0.1$, 0.03, and 0.01 and for SGLD at $h=0.1$, 0.03, and 0.01, against the push at different training steps. (a)--(f) Mutual response $\chi_{ij}(t,512,32)$ and (g)--(l) mutual response $\chi_{ij}(t,2,2)$. Pairs are grouped by distance. The vertical axes are logarithmic. Each point is the mean over the pushes made at one training step $t$, at the push steps that all three strengths share. All 512 networks of each optimizer are pushed. A network enters a point if at least one of its pushed copies passes the crash filter there at all three strengths. For AdamW these are all 512 networks at every point. For SGLD they are 224--387 of the 512 networks at $\tau=512$, and 486--504 at $\tau=2$. Bands show the 95\% bootstrap interval.}
    \label{fig:strength}
\end{figure*}

To check that the local interaction occurs in a broader range of push strengths, we repeated the experiments with pushes at $\alpha=0.03$ and 0.01 for AdamW and at $h=0.03$ and 0.01 for SGLD. Figure~\ref{fig:strength} shows the mutual response by distance for each strength. For both optimizers and at every strength, the response rises between steps 15\,000 and 40\,000 and shows the full order after step 40\,000 (for SGLD at $h=0.01$ from step 50\,000 on). Table~\ref{tab:strength} counts the networks with the full order. For SGLD the fraction is lower at $h=0.01$, 75\%, while neighbors still respond more than layers three apart in 97\% of the networks. This is because the push at $h=0.01$ is too weak to show the order in a single network. The response to one such push differs from push to push by 24 times the size of its mean. In the AdamW networks this difference is 0.4--0.6 times the size of the mean at every strength. However, the order is still present at $h=0.01$ [Fig.~\ref{fig:strength}(f)].

\begin{table*}[!t]
    \centering
    \caption{Fraction of networks with the order by distance, for each push strength, for $\chi_{ij}(t,512,32)$ and for $\chi_{ij}(t,2,2)$. The mutual response of a network is computed from its pushes made from step 60\,000 on, at the push steps that all three strengths share. For $\chi_{ij}(t,512,32)$, 510 AdamW and 276 SGLD networks enter, those that pass the crash filter at all three strengths. For $\chi_{ij}(t,2,2)$, all 512 networks of each optimizer enter. Fewer push steps enter than in Table~\ref{tab:fullorder}, and for $\chi_{ij}(t,512,32)$ not the same set of networks, so the fractions at $\alpha=0.1$ and $h=0.1$ differ from those there. The 95\% bootstrap interval is given in parentheses.}
    \label{tab:strength}
    \begin{tabular}{llcccc}
        \hline\hline
         & & \multicolumn{2}{c}{$\chi_{ij}(t,512,32)$} & \multicolumn{2}{c}{$\chi_{ij}(t,2,2)$} \\
        Method & Strength & Full order (\%) & Neighbors $>$ 3 apart (\%) & Full order (\%) & Neighbors $>$ 3 apart (\%) \\
        \hline
        AdamW & $\alpha=0.1$ & 99.4 (98.6--100.0) & 99.6 (99.0--100.0) & 95.9 (94.1--97.5) & 99.0 (98.0--99.8) \\
         & $\alpha=0.03$ & 100.0 (100.0--100.0) & 100.0 (100.0--100.0) & 98.0 (96.7--99.2) & 99.4 (98.6--100.0) \\
         & $\alpha=0.01$ & 99.8 (99.4--100.0) & 100.0 (100.0--100.0) & 98.0 (96.7--99.2) & 99.6 (99.0--100.0) \\
        \hline
        SGLD & $h=0.1$ & 93.5 (90.6--96.4) & 98.2 (96.4--99.6) & 69.3 (65.2--73.2) & 95.7 (93.9--97.3) \\
         & $h=0.03$ & 89.1 (85.5--92.4) & 98.6 (97.1--99.6) & 65.0 (60.7--69.1) & 94.9 (93.0--96.7) \\
         & $h=0.01$ & 75.0 (69.9--80.1) & 96.7 (94.6--98.6) & 59.6 (55.3--63.9) & 92.2 (89.8--94.3) \\
        \hline\hline
    \end{tabular}
\end{table*}

The local interaction is therefore not limited to the strengths $\alpha=0.1$ and $h=0.1$ used in the main text. Figure~\ref{fig:strength}(g)--(l) shows the same for the response one step after the first push. It hardly depends on the strength. From step 60\,000 on, the mutual response of neighbors is 0.11\% at every strength in the AdamW networks, and 0.014\%, 0.011\%, and 0.011\% at $h=0.1$, 0.03, and 0.01 in the SGLD networks. At every strength neighbors respond most and layers three apart least from step 60\,000 on. Counted in each network separately, the order at $\tau=2$ is less clear than at $\tau=512$. For SGLD the full order is found in 60--69\% of the networks at these push steps, and neighbors respond more than layers three apart in 92--96\%. One step after a single SGLD push, the change of the neighboring layer differs from push to push by 2.8--2.9 times its mean at every strength. The order is thus present from the first step on in the mean over the pushes, but it becomes stronger as the response accumulates over the steps.

\subsection{Distance and the overall strength of each layer}
\label{app:strengthfree}

The order of the averages in Table~\ref{tab:fullorder} could also come from the overall strength of layers that are included in many pairs rather than from the distance between them. The pairs two and three apart, 2--4, 3--5, and 2--5, all contain layer 2 or layer 5, so a weak layer 5 alone would make the averages fall with the distance. We assume that the overall strength of a layer enters the mutual response as a factor,
\begin{equation}
\chi_{ij}=G_i\,G_j\,g_{ij},
\label{eq:app-factor}
\end{equation}
where $\chi_{ij}$ is the mutual response of layers $i$ and $j$ in one network, computed as for Table~\ref{tab:fullorder}, $G_i$ belongs to layer $i$ alone, and $g_{ij}$ belongs to the pair and may depend on more than the distance. One step after the first push this form follows from Eq.~\eqref{eq:app-inner}, for SGLD at every step and for AdamW once $\sqrt{\hat v}$ is well below $\epsilon$. Then $G_i$ is proportional to the size of $\delta y_i$ divided by $\|\partial r_i/\partial W_i\|_F$, that is, to the change of the scores per unit size of the push, and $g_{ij}$ is the absolute value of the cosine between $\delta y_i$ and $\delta y_j$. At $\tau=512$ it is an assumption, since the response then also passes through the other layers. We remove the factors $G_i$ with two ratios in which each of layers 2--5 appears once above and once below,
\begin{equation}
C_2\equiv\left(\frac{\chi_{24}\,\chi_{35}}{\chi_{23}\,\chi_{45}}\right)^{1/2},\qquad C_3\equiv\frac{\chi_{25}\,\chi_{34}}{\chi_{23}\,\chi_{45}},
\label{eq:app-C}
\end{equation}
\begin{table*}[!t]
    \centering
    \caption{Ratios $C_2$ and $C_3$ of Eq.~\eqref{eq:app-C}. The filtered networks are those of Table~\ref{tab:fullorder}, except that for SGLD at $\tau=2$ the 124 networks (Early) and 4 networks (Late) in which one of the six pairs has a mutual response of zero are left out. The ratios are given as medians over the networks, because single networks reach ratios above 10. The 95\% bootstrap interval is given in parentheses below each value.}
    \label{tab:strengthfree}
    \begin{tabular}{lcccc}
        \hline\hline
         & \multicolumn{2}{c}{AdamW} & \multicolumn{2}{c}{SGLD} \\
         & Early & Late & Early & Late \\
        \hline
        \multicolumn{5}{l}{$\tau=512$} \\
        Median $C_2$ & 0.80 & 0.93 & 0.97 & 0.94 \\
         & (0.74--0.87) & (0.92--0.93) & (0.84--1.10) & (0.93--0.96) \\[3pt]
        Median $C_3$ & 0.71 & 0.87 & 1.00 & 0.99 \\
         & (0.63--0.82) & (0.86--0.88) & (0.70--1.25) & (0.92--1.04) \\[3pt]
        $C_2<1$ (\%) & 61.7 & 99.4 & 52.1 & 72.7 \\
         & (57.6--65.8) & (98.6--100.0) & (44.8--60.0) & (65.5--79.4) \\[3pt]
        $C_3<C_2<1$ (\%) & 34.2 & 74.8 & 27.3 & 33.9 \\
         & (30.1--38.3) & (71.1--78.5) & (20.6--33.9) & (26.7--41.2) \\[3pt]
        \multicolumn{5}{l}{$\tau=2$} \\
        Median $C_2$ & 0.88 & 0.99 & 0.86 & 0.64 \\
         & (0.79--0.94) & (0.99--0.99) & (0.77--0.91) & (0.59--0.69) \\[3pt]
        Median $C_3$ & 1.09 & 1.00 & 0.69 & 0.92 \\
         & (0.84--1.31) & (1.00--1.00) & (0.57--0.86) & (0.76--1.12) \\[3pt]
        $C_2<1$ (\%) & 57.6 & 79.7 & 60.1 & 72.8 \\
         & (53.3--61.9) & (76.2--83.2) & (55.2--64.9) & (69.1--76.6) \\[3pt]
        $C_3<C_2<1$ (\%) & 21.5 & 30.5 & 30.7 & 31.7 \\
         & (18.0--25.0) & (26.6--34.6) & (26.3--35.3) & (27.6--35.6) \\
        \hline\hline
    \end{tabular}
\end{table*}
so that $C_2=[g_{24}\,g_{35}/(g_{23}\,g_{45})]^{1/2}$ and $C_3=g_{25}\,g_{34}/(g_{23}\,g_{45})$. The ratio $C_2$ compares the pairs two apart with neighbors, and $C_3$ compares the pair three apart with neighbors. If $g_{ij}$ were the same for all pairs, both ratios would be 1, and if it depends only on the distance and falls with it, $C_2<1$ and $C_3<C_2$. The test can confirm an order but not rule one out. Under Eq.~\eqref{eq:app-factor} a weak layer cannot make $C_2$ or $C_3$ smaller than 1, so values below 1 show that the order is not produced by the strength of single layers. A value close to 1, however, does not show that the response is independent of the distance, because with four layers the two ratios compare only particular pairs, as discussed below for the pair 3--4.

With the strength of each layer removed, neighbors still respond more than layers two apart in most networks late in training (Table~\ref{tab:strengthfree}). At $\tau=512$, $C_2<1$ holds late in 99.4\% of the AdamW and in 72.7\% of the SGLD networks, against 61.7\% and 52.1\% early. For a response that does not depend on the distance it would hold in about half of the networks. In both methods the preference thus becomes more common during training, because $C_2$ varies less between the networks, not because it becomes smaller. The medians of $C_2$ and $C_3$ in the AdamW networks rise over time, from 0.80 and 0.71 early to 0.93 and 0.87 late. This rise does not mean that the preference becomes weaker, but that early the ratios differ so widely between the networks that their median does not describe a typical network. Early, many networks have a strong preference and many the opposite order, with the 10th and 90th percentiles over the AdamW networks of 0.30 and 2.1 for $C_2$ and of 0.13 and 4.5 for $C_3$. Late, these percentiles are close to 1 (0.89 and 0.98 for $C_2$, and 0.80 and 1.00 for $C_3$). In the SGLD networks the medians change little, and the percentiles of $C_2$ move from 0.36 and 2.5 early to 0.76 and 1.16 late.

One step after the first push, late in training, $C_2$ is close to 1 in the AdamW networks, although their response at this delay depends on the distance. At this delay the response of a pair follows the cosine between the changes of the scores of its two layers, $g_{ij}$ in Eq.~\eqref{eq:app-factor}, and this cosine is smaller for layers further apart (Table~\ref{tab:distance}). Each pair two apart in $C_2$, however, is obtained from a neighbor pair by replacing layer 3 with layer 4 or the reverse, 2--3 by 2--4 and 4--5 by 3--5. Such a replacement leaves the cosine unchanged if layers 3 and 4 change the scores in the same way, and in the AdamW networks they nearly do (cosine 0.99). The median $C_2$ of the AdamW networks is therefore 0.99. In the SGLD networks layers 3 and 4 differ (cosine 0.20), so the replacement lowers the cosine, and the median $C_2$ of the SGLD networks is 0.64. At $\tau=512$ the response also passes through the other layers, so $C_2$ is then no longer set by the one-step cosines alone.

Late in training, at $\tau=512$, whether layers three apart respond less than layers two apart differs between the methods. In the AdamW networks at $\tau=512$, the median $C_3=0.87$ is close to $C_2^2=0.86$ late in training. This is the relation expected when the pair factor of two layers is the product of those of the neighbor pairs between them, $g_{25}=g_{23}\,g_{34}\,g_{45}$, as for the cosines in Sec.~\ref{sec:why}.

With four layers to compare, however, this rule makes both ratios depend on the pair 3--4 alone, $C_2=g_{34}$ and $C_3=g_{34}^2$, since $g_{23}$ and $g_{45}$ cancel. The ratios then measure only what is lost between layers 3 and 4, not between layers 2 and 3 or 4 and 5. The explanation given above for the median $C_2$ of 0.99 in the AdamW networks one step after the first push therefore extends to $C_3$. With the cosine of 0.99 between layers 3 and 4, the rule gives $C_3=0.99^2\approx0.98$, close to the median $C_3$ of 1.00 at this delay. In the SGLD networks, late in training, the 95\% bootstrap interval of the median $C_3$ includes 1 at both delays, in line with their cosines of layers further apart, which are larger than the product of those of the neighbor pairs between them (Sec.~\ref{sec:why}). Whether the response keeps falling for layers more than two apart can therefore be tested only in deeper networks. We leave this test for future work.

The weak response of layer 5 steepens the fall of the averages with the distance, but it does not create this fall. The pairs that include layer 5 respond less than most other pairs at the same distance. Late in training, at $\tau=512$, with the pushes of all networks pooled as in Fig.~\ref{fig:adamw}, the neighbors 2--3, 3--4, and 4--5 have a mutual response of 6.93, 6.93, and 3.10\% in the AdamW networks and of 1.14, 2.30, and 1.08\% in the SGLD networks, and the pairs 2--4 and 3--5 have 6.42 and 2.89\% in the AdamW and 1.24 and 0.87\% in the SGLD networks. Among the pairs that include layer 5, however, the response still falls with the distance, from 3.10 to 2.89 and 2.65\% in the AdamW networks and from 1.08 to 0.87 and 0.50\% in the SGLD networks for the pairs 4--5, 3--5, and 2--5.

\subsection{Local interaction in networks that do not generalize}
\label{app:nogen}

The local interaction does not require a network to generalize. In 14 of the 512 AdamW networks the test accuracy never exceeds 0.03 during the 150\,000 training steps (a random guess gives $1/113\approx0.009$). Table~\ref{tab:nogen} compares them with the other 498 networks. Late in training their neighboring layers line up to 0.98, and all 14 are in the full order at $\tau=512$. At $\tau=2$ the full order holds in 11 of the 14, less often than in the other networks. In every SGLD network the test accuracy rises above 0.03, so this comparison cannot be made for SGLD.

\begin{table}[!t]
    \centering
    \caption{The 14 AdamW networks whose test accuracy never exceeds 0.03 and the other 498 AdamW networks. Early and Late are the pushes made up to step 15\,000 and from step 60\,000 on, as in Table~\ref{tab:fullorder}. The final test accuracy is the mean over the last 10\,000 training steps. The line-up is the mean over the three neighbor pairs, at step 10\,000 (Early) and over the steps from 60\,000 on (Late). The last two rows give the number of networks in the full order of $\chi_{ij}(t,512,32)$ ($\tau=512$) and of $\chi_{ij}(t,2,2)$ ($\tau=2$).}
    \label{tab:nogen}
    \begin{tabular}{lcccc}
        \hline\hline
         & \multicolumn{2}{c}{Never above 0.03} & \multicolumn{2}{c}{Other networks} \\
         & Early & Late & Early & Late \\
        \hline
        Networks & \multicolumn{2}{c}{14} & \multicolumn{2}{c}{498} \\
        Final test accuracy & & 0.02 & & 0.97 \\
        Line-up & 0.35 & 0.98 & 0.43 & 0.96 \\
        Full order, $\tau=512$ & 1 & 14 & 154 & 496 \\
        Full order, $\tau=2$ & 4 & 11 & 115 & 488 \\
        \hline\hline
    \end{tabular}
\end{table}

\section{Why layers that line up move the scores alike}
\label{app:alike}

Here we derive Eq.~\eqref{eq:alike}. We write $m_i(x)$ for the vector with the entries $m_{i,l}(x)$ of Eq.~\eqref{eq:balance-terms}, $\odot$ for the entry-by-entry multiplication of two vectors, and $\mathrm{diag}(m_i)$ for the diagonal matrix with the entries of $m_i$ on its diagonal. We also write $S_i(x)\equiv \partial y(x)/\partial z_i(x)$, with $z_5\equiv y$, so that $S_5$ is the identity matrix. Since $z_{i+1}=W_{i+1}(m_i\odot z_i)+b_{i+1}$, the matrix $S_i$ obeys $S_i=S_{i+1}W_{i+1}\,\mathrm{diag}(m_i)$. A small change $\Delta W_i$ of the weights changes $z_i$ by $\Delta W_ih_{i-1}$, and so it changes the scores by $S_i\,\Delta W_i\,h_{i-1}$, which is $J_i\Delta W_i$ in Eq.~\eqref{eq:routes}. For the direction $\partial r_i/\partial W_i$ of the push of Eq.~\eqref{eq:push}, we write the change of the scores per push size $\kappa$ as 
\begin{equation}
\delta y_i(x)\equiv S_i(x)\,\frac{\partial r_i}{\partial W_i}\,h_{i-1}(x).
\label{eq:app-dy}
\end{equation}

 The push on layer $i$ changes the scores by $\kappa\,\delta y_i$. After one step of size $\eta$, the response of layer $j$ from the first term of Eq.~\eqref{eq:routes} is
\begin{equation}
\Delta r_j^{(i)}(t,2) =-\frac{\eta\kappa}{N}\sum_x\delta y_j(x)^{\mathrm T}\,H(x)\,\delta y_i(x),\qquad H\equiv \frac{\partial q}{\partial y}.
\label{eq:app-inner}
\end{equation}
For the cross-entropy and for an arbitrary vector $\omega$,
\begin{equation}
\omega^{\mathrm T}H(x)\,\omega=\sum_kP_k\,\omega_k^2-\Bigl(\sum_kP_k\,\omega_k\Bigr)^2,
\end{equation}
where $P_k(x)\equiv e^{y_k(x)}/\sum_le^{y_l(x)}$ is the probability that the network gives to answer $k$. This is the variance of the numbers $\omega_k$ under these probabilities, so it is never negative. The sum in Eq.~\eqref{eq:app-inner} is therefore an inner product of the two changes of the scores. From the Cauchy--Schwarz inequality, it is largest when $\delta y_j(x)$ is $\delta y_i(x)$ times one positive factor that is the same for all examples.

Next we write the push with the patterns of the layer. Putting $W_i=\sum_k\sigma_ku_kv_k^{\mathrm T}$ into the gradient of Eq.~\eqref{eq:push} gives
\begin{equation}
\frac{\partial r_i}{\partial W_i}=\frac{2}{\sigma_1^2}\Bigl[-(r_i-1)\,\sigma_1u_1v_1^{\mathrm T}+\sum_{k\ge2}\sigma_ku_kv_k^{\mathrm T}\Bigr].
\label{eq:app-pushpatterns}
\end{equation}
To analyze the effect of line-up, let us keep only the term of the strongest pattern. By Eq.~\eqref{eq:app-dy} it changes the scores by
\begin{equation}
\delta y_i^{(1)}(x)=-\frac{2(r_i-1)}{\sigma_1(W_i)}\,S_i(x)\,u_1(W_i)\,\bigl[v_1(W_i)\cdot h_{i-1}(x)\bigr].
\label{eq:app-dy1}
\end{equation}

Now let layers $i$ and $i+1$ be lined up, $u_1(W_i)=v_1(W_{i+1})\equiv u$, and let the neurons on which $u$ is nonzero be turned on for every example, i.e., $m_i(x)\odot u=u$. Then, $S_iu=S_{i+1}W_{i+1}u=\sigma_1(W_{i+1})\,S_{i+1}u_1(W_{i+1})$, and $u\cdot h_i=u\cdot z_i=\sigma_1(W_i)\,[v_1(W_i)\cdot h_{i-1}]+u\cdot b_i$. With $a(x)\equiv v_1(W_i)\cdot h_{i-1}(x)$ and $w(x)\equiv S_{i+1}(x)\,u_1(W_{i+1})$, Eq.~\eqref{eq:app-dy1} becomes
\begin{align}
\delta y_i^{(1)}(x)&=-2(r_i-1)\,\frac{\sigma_1(W_{i+1})}{\sigma_1(W_i)}\,a(x)\,w(x),\label{eq:app-dyi}\\
\delta y_{i+1}^{(1)}(x)&=-2(r_{i+1}-1)\,\frac{\sigma_1(W_i)}{\sigma_1(W_{i+1})}\Bigl[a(x)+\frac{u\cdot b_i}{\sigma_1(W_i)}\Bigr]w(x).\label{eq:app-dyi1}
\end{align}
Both are multiples of the same vector $w(x)$. If the bias term is small compared with $a(x)$, they are proportional on every example, $\delta y_{i+1}^{(1)}(x)=c_i\,\delta y_i^{(1)}(x)$, with
\begin{equation}
c_i \approx \frac{(r_{i+1}-1)\,\sigma_1(W_i)^2}{(r_i-1)\,\sigma_1(W_{i+1})^2}>0.
\end{equation}
This is Eq.~\eqref{eq:alike} for the parts of the two pushes along the strongest patterns.

For layers that are further apart, Table~\ref{tab:distance} gives the cosine between the two changes of the scores, taken with the inner product of Eq.~\eqref{eq:app-inner} and averaged over the pairs of layers at the same distance, for both optimizers. In the AdamW networks the cosine falls with the distance from step 40\,000 on. In the SGLD networks the cosine is close to zero at step 20\,000. From step 40\,000 on it is largest for neighbors and falls with the distance. It stays smaller than in the AdamW networks, because the neighboring layers are only partly lined up due to the injected noise.

Only pushes on the strongest patterns make two layers change the scores alike. We repeated the calculation with two other pushes of each layer, one along a random pattern (a random output direction times a random input direction) and one along its tenth strongest pattern. From step 60\,000 on, the cosine between the changes of the scores made by two layers is then at most 0.03 in the AdamW networks and 0.01 in the SGLD networks for every pair, neighbors or not, against 0.77--0.99 and 0.02--0.20 for the push along the gradient of the stable rank, which moves mainly the strongest pattern of a layer. The cosine, and with it the order by distance, thus comes from the strongest patterns.

\begin{table}[!t]
    \caption{Cosine between the changes of the scores made by the pushes on two layers, in the AdamW and SGLD networks, averaged over the pairs of layers at the same distance. Means over the networks, at one training step or over all saved steps from 60\,000 on. The networks are the same as in Table~\ref{tab:routes}. The 95\% bootstrap interval is given in parentheses below each mean.}
    \label{tab:distance}
    \begin{tabular}{lccc}
        \hline\hline
        Training step & Neighbors & 2 apart & 3 apart \\
        \hline
        \multicolumn{4}{l}{AdamW} \\
        10\,000 & 0.19 & 0.22 & 0.24 \\
         & (0.17--0.20) & (0.20--0.24) & (0.21--0.27) \\[3pt]
        20\,000 & 0.41 & 0.49 & 0.61 \\
         & (0.39--0.43) & (0.47--0.51) & (0.58--0.64) \\[3pt]
        30\,000 & 0.75 & 0.66 & 0.67 \\
         & (0.73--0.77) & (0.63--0.68) & (0.64--0.70) \\[3pt]
        40\,000 & 0.84 & 0.78 & 0.68 \\
         & (0.82--0.85) & (0.76--0.80) & (0.66--0.71) \\[3pt]
        From 60\,000 & 0.91 & 0.87 & 0.77 \\
         & (0.91--0.92) & (0.86--0.88) & (0.75--0.78) \\
        \hline
        \multicolumn{4}{l}{SGLD} \\
        20\,000 & 0.00 & 0.00 & 0.01 \\
         & (0.00--0.01) & (0.00--0.00) & (0.00--0.01) \\[3pt]
        40\,000 & 0.17 & 0.12 & 0.08 \\
         & (0.16--0.19) & (0.11--0.13) & (0.07--0.08) \\[3pt]
        From 60\,000 & 0.12 & 0.04 & 0.02 \\
         & (0.11--0.12) & (0.04--0.04) & (0.01--0.02) \\
        \hline\hline
    \end{tabular}
\end{table}

The derivation required the neurons on which $u$ is nonzero to be on. This is close to the condition of Appendix~\ref{app:offshare}, a small off-ratio, which is defined there and measured for the SGLD networks to be small on average. Also, Eq.~\eqref{eq:app-inner} is for a gradient step. An AdamW step divides the mean gradient of each weight by its own $\sqrt{\hat v}+\epsilon$, so Eq.~\eqref{eq:app-inner} holds up to a constant only once $\sqrt{\hat v}$ has fallen well below $\epsilon$. The median of $\sqrt{\hat v}/\epsilon$ over the weights of a layer is 117--214 at step 10\,000, 0.8--1.5 at step 20\,000, and 0.02--0.07 from step 30\,000 on. 

\section{Robustness of the line-up--response relation across individual layer pairs}
\label{app:pairs}

The main text shows the line-up and the mutual response of neighbors as means over the three neighbor pairs (Figs.~\ref{fig:adamw} and~\ref{fig:lineup}). Here we show the line-up, the response, and the correlation between the two for each neighbor pair by itself.

\begin{figure*}[!t]
    \centering
    \includegraphics[width=\textwidth]{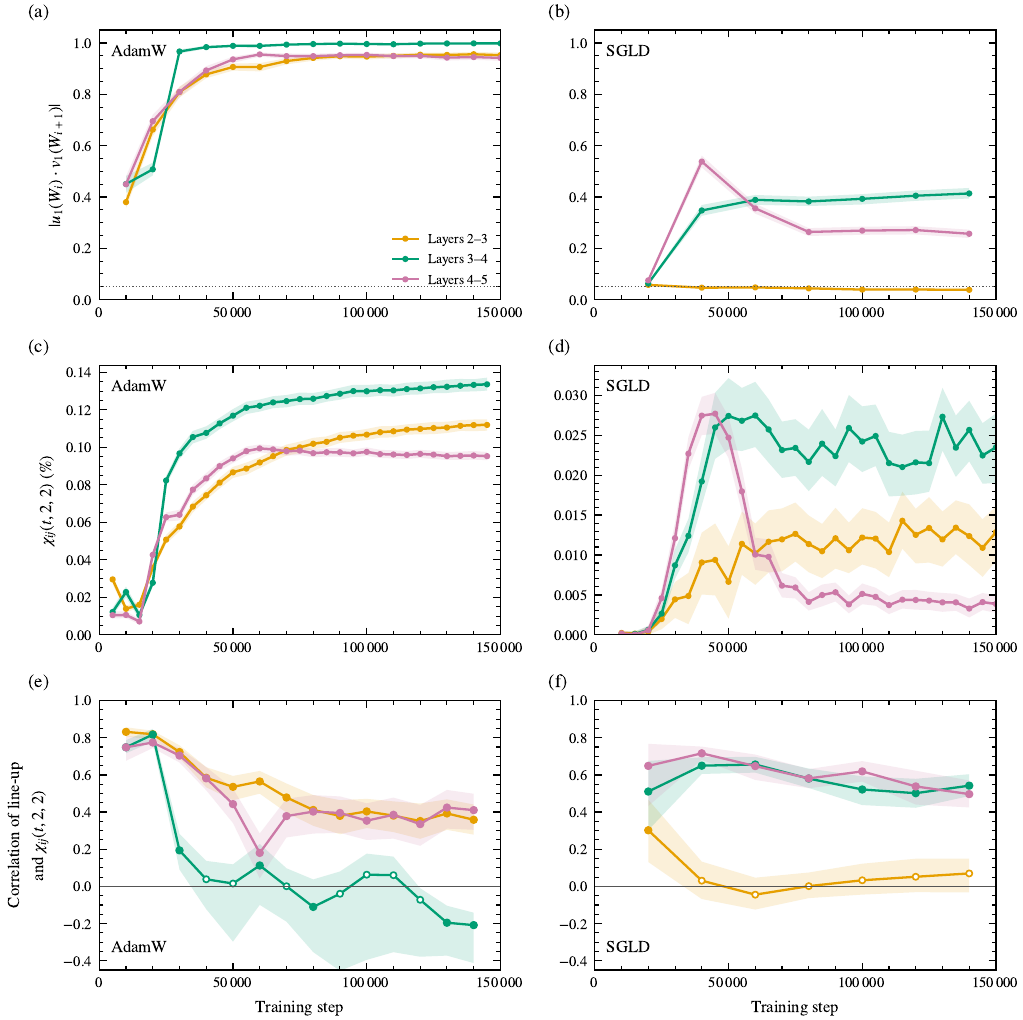}
    \caption{Line-up and response of each neighbor pair, and their correlation, for AdamW (left) and SGLD (right). (a), (b) Line-up of each pair against the training step, as the mean over the networks. The dotted line marks the level of unrelated patterns (0.050). (c), (d) Mutual response one step after the first push, $\chi_{ij}(t,2,2)$. (e), (f) Correlation across the networks between the line-up of a pair and $\chi_{ij}(t,2,2)$. Open symbols mark correlations that are not significant in a permutation test, with the procedure of Ref.~\cite{benjamini1995controlling} for many tests, set to a false discovery rate of 5\%. All panels use the networks that pass the crash filter, all 512 AdamW networks and 479--507 SGLD networks at each point. Bands show the 95\% bootstrap interval.}
    \label{fig:pairs}
\end{figure*}

Figure~\ref{fig:pairs}(a), (b) shows the line-up of each neighbor pair, and Fig.~\ref{fig:pairs}(c), (d) its mutual response one step after the first push. In the AdamW networks every pair lines up between steps 10\,000 and 30\,000 (0.38--0.45 at step 10\,000 and 0.81--0.97 at step 30\,000), and the response of every pair rises there. In the SGLD networks layers 3--4 and 4--5 are partly lined up at step 40\,000 (0.35 and 0.54), whereas layers 2--3 stay at the level of unrelated patterns at every step (0.04--0.06). Layers 4--5 have the largest response of the three pairs for pushes made from step 25\,000 to step 45\,000, and the smallest from step 65\,000 on, while their line-up falls. Layers 2--3 respond least for pushes made from step 20\,000 to step 55\,000.

The levels of different pairs cannot be compared by their line-up alone. In the SGLD networks the same line-up gives a different size of the response in different pairs, since a push changes the scores by a different amount in each layer, about twice as much for layer 2 as for layer 5 from step 60\,000 on. In the AdamW networks this amount differs by less than 10\% between the layers, and there the pair with the largest line-up also has the largest response one step after the push, from step 20\,000 on. Single pairs need not follow the order by distance exactly. For example, layers 2 and 3 of the SGLD networks do not line up. The order is a property of the mean over the pairs at each distance, and late in training this mean falls with the distance in 97.5--99.6\% of the AdamW and 82.4--92.1\% of the SGLD networks (Table~\ref{tab:fullorder}).

The two strongest patterns of a layer can have nearly equal strengths. From step 60\,000 on, $\sigma_1/\sigma_2$ is 1.03--1.04 in every layer from 2 to 5, except layer 5 of the SGLD networks with 1.11 (means over the 512 networks of each method), so the strongest pattern is not far stronger than the second. Which of the two is the strongest can then depend on small differences. Two layers can share the same two patterns but have a different one as their strongest. So the line-up, which compares only the strongest patterns, can then be low although the patterns are shared. We therefore also compare the five strongest patterns of the two layers as a group, which does not depend on their order. This is the measure of Eq.~\eqref{eq:app-partial} with $d=5$, a mean squared overlap, which is 1 when the two groups are the same and $5/256\approx0.020$ for unrelated groups. From step 60\,000 on it is 0.98--1.00 for the AdamW pairs, and 0.48 and 0.40 for the SGLD pairs 3--4 and 4--5, whereas for layers 2--3 it is 0.016, at the level of unrelated patterns. The near-equal strengths thus do not change the conclusions drawn from the strongest patterns.

Figure~\ref{fig:pairs}(e), (f) shows, at each training step, the correlation between the response and the line-up, which Sec.~\ref{sec:why} gives only as one range of values, to test whether the relation holds throughout training. At each step, it is the Pearson correlation coefficient across the networks between the line-up of a pair and its mutual response one step after the first push. It is 1 when the response rises linearly with the line-up, and 0 when the two are unrelated.

The correlation is positive for every pair that lines up, and close to zero from step 40\,000 on for layers 2 and 3 of the SGLD networks, the pair that does not line up. Over all steps it is 0.18--0.83 for the pairs that line up, with one exception, and between $-0.04$ and 0.30 for layers 2 and 3 of the SGLD networks. The exception is layers 3--4 of the AdamW networks, whose correlation falls after step 20\,000 to about zero, and to $-0.21$ at step 140\,000. This drop need not mean that the relation between the line-up and the response is weaker. From step 30\,000 on, these two layers are lined up almost equally in every network (mean line-up 0.97--1.00, with the 10th and 90th percentiles over the networks within 0.06 of each other), so their line-up cannot account for the differences between the responses of the networks.

We also checked that the line-up and the response are not correlated merely because both depend on a common quantity. Five possible candidates are the stable ranks of the two layers, which enter both the size of the push and the factor $c_i$ of Eq.~\eqref{eq:alike}, as well as the Frobenius norms of the two layers and the logarithm of the training loss (because it differs by many factors of 10 between networks). At each training step, let $Y_n$ be the line-up or the response of network $n$, and let the five candidates be $X_{n,1},\dots,X_{n,5}$. We fit $Y_n$ by a linear function of these numbers and keep what the fit leaves over,
\begin{equation}
\zeta_n\equiv Y_n-\nu_0-\sum_{k=1}^{5}\nu_kX_{n,k},
\end{equation}
where the coefficients $\nu_0,\dots,\nu_5$ are chosen so that $\sum_n\zeta_n^2$ is smallest. We then correlate the leftovers $\zeta_n$ of the line-up and of the response across the networks. This correlation would be zero if the line-up and the response were linked only through the five numbers. It is 0.51--0.88 for the pairs that line up, with the same exception, close to the values above.

\section{Derivation of the line-up between neighboring layers}
\label{app:lineup}

This appendix demonstrates the theoretical background of Sec.~\ref{sec:balance}. We derive which neighbor pairs line up and how much of the line-up stays visible under the SGLD noise. For this we derive how training changes the difference $D_i$ in ReLU layers with biases, weight decay, steps of finite size, and the SGLD noise together. Our equation generalizes results for narrower cases, which are the one-step equation for $D_i$ without the ReLU~\cite{wang2024implicit} and for a single neuron~\cite{kunin2021neural}, the shrinking of $D_i$ by weight decay without an activation function~\cite{wang2024implicit,kobayashi2024weight,jacot2025wide}, the form of $K_i$ for smooth activations~\cite{rawal2026saddle}, and its zero entries for the ReLU~\cite{du2018algorithmic,le2022training}.

\subsection{Notation and definitions}
\label{app:quantity}

The main text states that two neighboring layers are lined up if $A_i=B_i$. Here we make this statement exact. The input patterns $v_k$ of a matrix $W$ are the eigenvectors of $W^{\mathrm T} W$, and its output patterns $u_k$ are the eigenvectors of $WW^{\mathrm T}$, both with eigenvalues $\sigma_k^2$. The matrix $B_i=W_iW_i^{\mathrm T}+b_ib_i^{\mathrm T}$ contains the bias of layer $i$. The line-up in the figures is measured with $W_i$ without the bias. In our experiments, the bias adds at most 0.7\% of $\sigma_1^2$ to $B_i$ along the strongest pattern, so this makes little difference.
If $D_i=0$, the two layers have the same nonzero singular values. If, in addition, the nonzero eigenvalues of $A_i$ are nondegenerate, the input patterns of $W_{i+1}$ with nonzero strength are exactly the output patterns of $\overline W_i$ with the same strength, up to sign. The question is therefore what training does to $D_i$.

In this appendix the function $\varphi$ in $h_i=\varphi(z_i)$ can be any activation function, not only the ReLU. The vector $e_i\equiv\partial\ell/\partial h_i$ has the entries $e_{i,k}$, and the factor of the noise in an SGLD step is written $s\equiv \sqrt{2\eta T}$.

\subsection{How training changes the difference}
\label{app:training}

This subsection derives Eq.~\eqref{eq:balance} and the entries of $K_i$ in Eq.~\eqref{eq:balance-terms}. Without its noise term $s\xi$, the SGLD step from $t$ to $t+1$ gives the weights $M_{i+1}=(1-\eta\lambda)W_{i+1}-\eta g_{i+1}$ and $\overline M_i=(1-\eta\lambda)\overline W_i-\eta\overline g_i$. The weights and gradients on the right-hand sides are those at step $t$. Multiplying out,
\begin{equation}
\begin{aligned}
M_{i+1}^{\mathrm T} M_{i+1}-\overline M_i\overline M_i^{\mathrm T}&=(1-\eta\lambda)^2D_i(t)+\eta^2Q_i\\
&\quad-\eta(1-\eta\lambda)(K_i+K_i^{\mathrm T}),
\end{aligned}
\label{eq:app-noiseless}
\end{equation}
with
\begin{equation}
K_i\equiv W_{i+1}^{\mathrm T} g_{i+1}-\overline g_i\overline W_i^{\mathrm T},\qquad Q_i= g_{i+1}^{\mathrm T} g_{i+1}-\overline g_i\overline g_i^{\mathrm T}.
\label{eq:app-KQ}
\end{equation}
The noise adds $s\xi_{i+1}$ and $s\overline\xi_i$, so $W_{i+1}(t+1)=M_{i+1}+s\xi_{i+1}$ and $\overline W_i(t+1)=\overline M_i+s\overline\xi_i$. Multiplying out $A_i(t+1)-B_i(t+1)$ then gives, for any step size $\eta$, exactly
\begin{equation}
\begin{aligned}
D_i(t+1)&=(1-\eta\lambda)^2D_i(t)-\eta(1-\eta\lambda)(K_i+K_i^{\mathrm T})\\
&\quad+\eta^2Q_i+R_i,
\end{aligned}
\label{eq:app-onestep}
\end{equation}
where
\begin{equation}
\begin{aligned}
R_i&= s\,(M_{i+1}^{\mathrm T}\xi_{i+1}+\xi_{i+1}^{\mathrm T} M_{i+1})-s\,(\overline M_i\overline\xi_i^{\mathrm T}+\overline\xi_i\overline M_i^{\mathrm T})\\
&\quad+s^2(\xi_{i+1}^{\mathrm T}\xi_{i+1}-\overline\xi_i\overline\xi_i^{\mathrm T}).
\end{aligned}
\label{eq:app-R}
\end{equation}
The noise is drawn independently of the weights, and its numbers have mean 0 and variance 1. So the mean of $\xi_{i+1}^{\mathrm T}\xi_{i+1}$ is $n_{i+1}I$, where $I$ is the identity matrix and $n_i$ is the number of neurons of layer $i$. In the same way the mean of $\overline\xi_i\overline\xi_i^{\mathrm T}$ is $(n_{i-1}+1)I$. So the mean of $R_i$ over the noise is $s^2(n_{i+1}-n_{i-1}-1)\,I$. Without weight decay, noise of this kind makes such differences grow without limit, as was found in a simulation of two weights for which multiplying one by a number and dividing the other by it leaves the network unchanged~\cite{ziyin2025noise}. Here weight decay keeps the mean of $D_i$ bounded.

For layers without the ReLU, it is known that weight decay shrinks this difference~\cite{wang2024implicit,kobayashi2024weight,jacot2025wide}. In the following, we show that it stays bounded with an activation function. An earlier bound~\cite{nystrom2026commutator} covers networks without biases, in the limit of small steps. Ours includes the biases and holds for steps of finite size. Without the noise, using Eq.~\eqref{eq:app-onestep} for $t$ steps gives
\begin{equation}
\begin{aligned}
\|D_i(t)\|_{F}&\le(1-\eta\lambda)^{2t}\|D_i(0)\|_{F}\\
&\quad+\frac{2(1-\eta\lambda)\max\|K_i\|_{F}+\eta\max\|Q_i\|_{F}}{\lambda(2-\eta\lambda)},
\end{aligned}
\label{eq:app-bound}
\end{equation}
where the maxima are over the previous $t$ steps. The first term falls by a factor $e$ in $1/(2\eta\lambda)=5000$ steps here, so the initial difference is lost in the large-$t$ limit. Then $\|D_i(t)\|_{F}$ is at most about $(\max\|K_i\|_{F}+\frac\eta2\max\|Q_i\|_{F})/\lambda$. In the limit $\eta\to0$ the part with $Q_i$ vanishes and only the part with $K_i$ stays, so the bound becomes $\max\|K_i\|_F/\lambda$. So the difference that weight decay leaves is set by $K_i$ and by the finite step size. 

The matrix $K_i$ can be written with the activation function. Write $\psi_{i+1}\equiv \partial\ell/\partial z_{i+1}$. Then the gradient is $g_{i+1}=\frac1N\sum_x\psi_{i+1}h_i^{\mathrm T}$, and $e_i=W_{i+1}^{\mathrm T}\psi_{i+1}$. Each output $h_{i,k}=\varphi(z_{i,k})$ depends only on its own input, so $\partial\ell/\partial z_i=\varphi'(z_i)\odot e_i$. In the same way as for $g_{i+1}$, the gradient is $\overline g_i=\frac1N\sum_x(\varphi'(z_i)\odot e_i)\,\overline h_{i-1}^{\mathrm T}$, where $\overline h_{i-1}$ is $h_{i-1}$ with one more entry 1. Hence
\begin{align}
W_{i+1}^{\mathrm T} g_{i+1}&=\frac1N\sum_x e_i\,\varphi(z_i)^{\mathrm T},\label{eq:app-grad1}\\
\overline g_i\,\overline W_i^{\mathrm T}&=\frac1N\sum_x(\varphi'(z_i)\odot e_i)\,(\overline W_i\,\overline h_{i-1})^{\mathrm T}\nonumber\\
&=\frac1N\sum_x(\varphi'(z_i)\odot e_i)\,z_i^{\mathrm T}.\label{eq:app-grad2}
\end{align}
Subtracting Eq.~\eqref{eq:app-grad2} from Eq.~\eqref{eq:app-grad1} gives, for any activation function $\varphi$,
\begin{equation}
(K_i)_{kl}=\frac1N\sum_x e_{i,k}(x)\,\bigl(\varphi(z_{i,l}(x))-\varphi'(z_{i,k}(x))\,z_{i,l}(x)\bigr).
\label{eq:app-K}
\end{equation}
If $\varphi(z)=\varphi'(z)\,z$ for every $z$, as for the ReLU, the leaky ReLU, and the identity, this becomes
\begin{equation}
(K_i)_{kl}=\frac1N\sum_x e_{i,k}(x)\,z_{i,l}(x)\,\bigl(\varphi'(z_{i,l}(x))-\varphi'(z_{i,k}(x))\bigr).
\label{eq:app-Krelu}
\end{equation}
Then the diagonal of $K_i$ is zero. An entry off the diagonal, $(K_i)_{kl}$, is zero when neurons $k$ and $l$ have the same slope on every example. Without an activation function all slopes are equal, so $K_i=0$. For the ReLU the slope $\varphi'(z_{i,l})$ is the switch $m_{i,l}$, so this is the formula of Eq.~\eqref{eq:balance-terms}, and $(K_i)_{kl}=0$ when the two neurons are on together or off together on every example. For a smooth activation such as $\tanh$, $\varphi(z)\neq\varphi'(z)\,z$, and even the diagonal is in general not zero. The zero diagonal is a known result for the ReLU, the leaky ReLU, and the identity~\cite{du2018algorithmic}. For ReLU networks without biases and with one output, trained with very small steps while no weight changes sign, it was shown that the entries of $K_i$ vanish for two neurons that switch together, and that the difference $D_i$ as a whole is not kept fixed in general~\cite{le2022training}. 

We now find the difference $D_i$ where training stops. Let $G_{i+1}\equiv g_{i+1}+\lambda W_{i+1}$ and $\overline G_i\equiv\overline g_i+\lambda\overline W_i$ be the gradients of the loss plus weight decay, $L+\frac\lambda2\sum_\theta\theta^2$, for these weights. Without the noise, a step changes the weights by $-\eta$ times these gradients, so training stops where they are zero. Putting $g_{i+1}=G_{i+1}-\lambda W_{i+1}$ and $\overline g_i=\overline G_i-\lambda\overline W_i$ into the definition of $K_i$ gives, at every value of the weights,
\begin{equation}
K_i+\lambda D_i=W_{i+1}^{\mathrm T} G_{i+1}-\overline G_i\overline W_i^{\mathrm T} .
\label{eq:app-stop}
\end{equation}
So where these gradients are zero, $D_i=-K_i/\lambda$ exactly. For any activation function this relation also follows from the identities for one layer in Ref.~\cite{ziyin2025formation}, applied to $W_{i+1}$ and to $\overline W_i$. If moreover $K_i=0$, as without an activation function, then $D_i=0$, and by Appendix~\ref{app:quantity} the two layers are exactly lined up.

\subsection{The pair that does not line up}
\label{app:offshare}

This subsection gives the measurement behind the statement of the main text on layers 2--3 of the SGLD networks, the pair that does not line up [Fig.~\ref{fig:pairs}(b)]. Where training without the noise stops, $D_i=-K_i/\lambda$, and by Eq.~\eqref{eq:app-Krelu} an entry of $K_i$ is not zero only if one of its two neurons is on and the other is off on some example. So the two layers agree on every pattern $u$ made of neurons that are always on, i.e., $u^{\mathrm T}D_iu=0$. They need not line up on a pattern that includes neurons that are off on some examples.

For the strongest output pattern $u_1$ of $W_i$, we measure this by the off-ratio
\begin{equation}
f_i\equiv\frac{\sum_x\bigl[u_1\cdot\bigl((1-m_i)\odot z_i\bigr)\bigr]^2}{\sum_x(u_1\cdot z_i)^2},
\label{eq:app-off}
\end{equation}
where $(1-m_i)\odot z_i$ is $z_i$ with the neurons that are on set to zero. Table~\ref{tab:offshare} shows the result for the SGLD networks from step 60\,000 on. For layer 2 the off-ratio is large, $f_2=0.57$, whereas it is 0.05 and 0.18 for layers 3 and 4. Layers 2--3 are the pair that does not line up. The off-ratio thus decides whether a pair can line up.

However, within one pair of layers, the off-ratio hardly sets how strongly each network lines up. The reason is the form of the rule above. The rule guarantees that the two layers agree on a pattern only if the pattern is made of neurons that are always on. For a pattern that includes neurons that are off on some examples, the rule allows the two layers to differ by an amount set by the entries of $K_i$, which are not related to the off-ratio. Among networks whose off-ratio is small, how strongly a pair lines up is then set by other quantities, for example how clearly the pattern stands out against the noise (Appendix~\ref{app:noise}).

The correlations across the networks of one pair agree with this (Table~\ref{tab:offshare}). In layers 3--4 the off-ratio is small (0.05), and the line-up varies only weakly with it (Pearson correlation $-0.28$) but clearly with the strength $\sigma_1$ of the strongest pattern of layer 3 (0.57). In layers 2--3 the off-ratio is large (0.57), and the pair does not line up (0.04). In layers 4--5 the off-ratio is larger than in layers 3--4 (0.18 against 0.05), so more of the pattern includes neurons that are off, and networks with a larger off-ratio line up less ($-0.37$), as the rule allows. The line-up there does not follow $\sigma_1$ ($-0.18$), because in layer 4 the networks with larger $\sigma_1$ also have a larger off-ratio (correlation 0.77 across the networks), which lowers their line-up. With the off-ratio held fixed, the line-up rises with $\sigma_1$ (0.17), as in layers 3--4. In the AdamW networks the off-ratio of Eq.~\eqref{eq:app-off} computed for the projection $P^{\mathrm{out}}_i$ onto the five strongest output patterns [Eq.~\eqref{eq:app-proj} with $d=5$] in place of $u_1$ is on average below 0.005 for all three pairs, and all three line up.

\begin{table*}[!t]
    \caption{Mean off-ratio $f_i$ of the strongest output pattern of $W_i$ and mean line-up of layers $i$ and $i+1$, for the neighbor pairs of the SGLD networks from step 60\,000 on. The last two columns give the Pearson correlation across the networks of the line-up with $f_i$ and with the strength $\sigma_1$ of the strongest pattern of $W_i$. All 512 SGLD networks enter. The 95\% bootstrap interval is given in parentheses below each value.}
    \label{tab:offshare}
    \begin{tabular}{lcccc}
        \hline\hline
        Layers & Off-ratio & Line-up & \multicolumn{2}{c}{Correlation with} \\
         & $f_i$ & & $f_i$ & $\sigma_1$ \\
        \hline
        2--3 & 0.57 & 0.04 & 0.09 & $-0.06$ \\
         & (0.56--0.58) & (0.04--0.04) & ($-0.06$ to 0.28) & ($-0.16$ to 0.02) \\[3pt]
        3--4 & 0.05 & 0.40 & $-0.28$ & 0.57 \\
         & (0.04--0.06) & (0.39--0.41) & ($-0.36$ to $-0.16$) & (0.51--0.63) \\[3pt]
        4--5 & 0.18 & 0.28 & $-0.37$ & $-0.18$ \\
         & (0.18--0.19) & (0.27--0.28) & ($-0.44$ to $-0.29$) & ($-0.27$ to $-0.09$) \\
        \hline\hline
    \end{tabular}
\end{table*}

\subsection{Why the line-up is only partial}
\label{app:noise}

The noise that SGLD adds to the weights can explain why the line-up of the SGLD networks is only partial and appears only after step 20\,000, as stated in the main text. This noise changes the weights at every step, so training never stops and $D_i$ is not zero. From step 60\,000 on, the Frobenius norm of $D_i$ restricted to the five strongest output patterns of $W_i$ is 0.7--1.7 times $\sigma_1(W_i)^2$. A pattern that a layer has learned is not visible if it is weaker than the noise. Once a learned pattern is stronger than the noise, it is among the strongest patterns but is mixed with noise. The measured line-up therefore stays below 1 even when the learned parts of the two layers are exactly lined up. The model below quantifies these statements.

To see what the noise does, we model each layer as a learned part $W^\star$ plus a noise matrix $E$,
\begin{equation}
W=W^\star+E,\qquad \sigma_E=\sqrt{\frac{2T}{\lambda(2-\eta\lambda)}},
\label{eq:app-model}
\end{equation}
where $E$ has independent Gaussian entries of mean 0 and standard deviation $\sigma_E$. This is the stationary standard deviation of a weight under weight decay and the noise alone. We also assume that the learned parts are balanced,
\begin{equation}
W^{\star\mathrm T}_{i+1}W^\star_{i+1}=W^\star_iW^{\star\mathrm T}_i,
\label{eq:app-balanced}
\end{equation}
so that without the noise the two layers would be exactly lined up.

We now quantify how much the noise reduces the line-up. The strength of the $j$th learned pattern relative to the noise is
\begin{equation}
\rho_j\equiv\frac{\sigma_j(W^\star_i)}{\sigma_E\sqrt n},
\label{eq:app-rho}
\end{equation}
where $n$ is the number of neurons of the layer. Two known results apply to this model in the limit of wide layers with a fixed number of learned patterns and fixed, different $\rho_j$. 

First, for a square $W_i$, a learned pattern is found among the strongest patterns of $W_i$ only if $\rho_j>1$. For one learned pattern, this was proven for the strongest pattern of $W_i$~\cite{benaychgeorges2012singular}. For several learned patterns, it was proven for any fixed number of the strongest patterns~\cite{ding2020high}. The limits for $\rho_j>1$ below were proven for several learned patterns~\cite{benaychgeorges2012singular}. If $\rho_j\le1$, the overlap $|u_k(W_i)\cdot u_j(W^\star_i)|$ of the $j$th learned pattern with the $k$th pattern of $W_i$ tends to zero for every fixed $k$. If $\rho_j>1$, the $j$th singular value of $W_i$ rises above $2\sigma_E\sqrt n$, the largest singular value of the noise alone, and the squared overlap $\mu_j^2$ of the observed pattern with the learned one has a nonzero limit,
\begin{align}
&\sigma_j(W_i)\to\sigma_E\sqrt n\,\Bigl(\rho_j+\frac1{\rho_j}\Bigr),\label{eq:app-sv}\\
&\mu_j^2\equiv|u_j(W_i)\cdot u_j(W^\star_i)|^2\to1-\frac1{\rho_j^2}.\label{eq:app-a}
\end{align}
A learned pattern is therefore visible in $W_i$ only once its strength exceeds $\sigma_E\sqrt n$, and the observed pattern approaches it as the strength grows. Before that, line-up cannot be measured.

Second, the measured line-up is partial. Let
\begin{equation}
P^{\mathrm{out}}_i\equiv\sum_{j=1}^{d}u_ju_j^{\mathrm T},\qquad P^{\mathrm{in}}_i\equiv\sum_{j=1}^{d}v_jv_j^{\mathrm T}
\label{eq:app-proj}
\end{equation}
be the projection operators onto the spaces spanned by the $d$ strongest output patterns $u_j$ and input patterns $v_j$ of $W_i$. In the limit of wide layers, the squared line-up of layers $i$ and $i+1$, which is $|u_1(W_i)\cdot v_1(W_{i+1})|^2$ for $d=1$, approaches the mean of the products $\mu_j^2\gamma_j^2$,
\begin{equation}
\frac1{d}\,\bigl\|P^{\mathrm{in}}_{i+1}P^{\mathrm{out}}_i\bigr\|_F^2\to\frac1{d}\sum_{j=1}^{d}\mu_j^2\gamma_j^2,
\label{eq:app-partial}
\end{equation}
where $\gamma_j^2\equiv|v_j(W_{i+1})\cdot v_j(W^\star_{i+1})|^2$ is the squared overlap for the input pattern of $W_{i+1}$. By Eq.~\eqref{eq:app-balanced} the learned parts of the two layers have the same strengths, so for square layers $\gamma_j^2$ has the same limit as $\mu_j^2$, and the right-hand side becomes $\frac1{d}\sum_{j=1}^{d}(1-1/\rho_j^2)^2$ when all $\rho_j>1$. Equation~\eqref{eq:app-partial} is a known result for two matrices that share a pattern and have independent noise. It was found for one shared pattern of symmetric matrices~\cite{pacco2023overlaps}. For rectangular matrices it was proven for one shared pattern and extended to several shared patterns of distinct strengths~\cite{baharav2025stacked}. It applies here because the learned parts of two neighboring layers share their patterns by Eq.~\eqref{eq:app-balanced}, while the noise of one layer is independent of the noise of the other. Both $\mu_j^2$ and $\gamma_j^2$ are below 1, so the measured line-up is below 1 even though the learned parts are exactly lined up.

In our layers of 256 neurons, Eqs.~\eqref{eq:app-sv}--\eqref{eq:app-partial} hold only approximately. However, their prediction that the line-up can appear only once the learned patterns are stronger than the noise still holds. In the SGLD networks, layers 2--5 all have a singular value above the largest singular value of the noise alone in 0.2\% of the 507 networks at step 20\,000 and in 92\% of the 491 networks at step 40\,000. The line-up of layers 3--4 and 4--5 first appears in the same range of steps, rising from 0.06 and 0.08 to 0.35 and 0.54. Layers 2--3 stay at the level of unrelated patterns (0.04--0.06), although their learned patterns are also stronger than the noise from step 40\,000 on. Patterns stronger than the noise are thus needed for a line-up but not enough, and Appendix~\ref{app:offshare} shows why layers 2--3 do not line up.

\bibliography{ref}

\begin{thebibliography}{57}%
\makeatletter
\providecommand \@ifxundefined [1]{%
 \@ifx{#1\undefined}
}%
\providecommand \@ifnum [1]{%
 \ifnum #1\expandafter \@firstoftwo
 \else \expandafter \@secondoftwo
 \fi
}%
\providecommand \@ifx [1]{%
 \ifx #1\expandafter \@firstoftwo
 \else \expandafter \@secondoftwo
 \fi
}%
\providecommand \natexlab [1]{#1}%
\providecommand \enquote  [1]{``#1''}%
\providecommand \bibnamefont  [1]{#1}%
\providecommand \bibfnamefont [1]{#1}%
\providecommand \citenamefont [1]{#1}%
\providecommand \href@noop [0]{\@secondoftwo}%
\providecommand \href [0]{\begingroup \@sanitize@url \@href}%
\providecommand \@href[1]{\@@startlink{#1}\@@href}%
\providecommand \@@href[1]{\endgroup#1\@@endlink}%
\providecommand \@sanitize@url [0]{\catcode `\\12\catcode `\$12\catcode
  `\&12\catcode `\#12\catcode `\^12\catcode `\_12\catcode `\%12\relax}%
\providecommand \@@startlink[1]{}%
\providecommand \@@endlink[0]{}%
\providecommand \url  [0]{\begingroup\@sanitize@url \@url }%
\providecommand \@url [1]{\endgroup\@href {#1}{\urlprefix }}%
\providecommand \urlprefix  [0]{URL }%
\providecommand \Eprint [0]{\href }%
\providecommand \doibase [0]{https://doi.org/}%
\providecommand \selectlanguage [0]{\@gobble}%
\providecommand \bibinfo  [0]{\@secondoftwo}%
\providecommand \bibfield  [0]{\@secondoftwo}%
\providecommand \translation [1]{[#1]}%
\providecommand \BibitemOpen [0]{}%
\providecommand \bibitemStop [0]{}%
\providecommand \bibitemNoStop [0]{.\EOS\space}%
\providecommand \EOS [0]{\spacefactor3000\relax}%
\providecommand \BibitemShut  [1]{\csname bibitem#1\endcsname}%
\let\auto@bib@innerbib\@empty
\bibitem [{\citenamefont {Zeiler}\ and\ \citenamefont
  {Fergus}(2014)}]{zeiler2014visualizing}%
  \BibitemOpen
  \bibfield  {author} {\bibinfo {author} {\bibfnamefont {M.~D.}\ \bibnamefont
  {Zeiler}}\ and\ \bibinfo {author} {\bibfnamefont {R.}~\bibnamefont
  {Fergus}},\ }\bibfield  {title} {\bibinfo {title} {Visualizing and
  understanding convolutional networks},\ }in\ \href@noop {} {\emph {\bibinfo
  {booktitle} {Computer Vision -- {ECCV} 2014}}},\ \bibinfo {series} {Lecture
  Notes in Computer Science}, Vol.\ \bibinfo {volume} {8689}\ (\bibinfo
  {publisher} {Springer},\ \bibinfo {year} {2014})\ pp.\ \bibinfo {pages}
  {818--833},\ \Eprint {https://arxiv.org/abs/1311.2901} {arXiv:1311.2901}
  \BibitemShut {NoStop}%
\bibitem [{\citenamefont {Yosinski}\ \emph {et~al.}(2014)\citenamefont
  {Yosinski}, \citenamefont {Clune}, \citenamefont {Bengio},\ and\
  \citenamefont {Lipson}}]{yosinski2014transferable}%
  \BibitemOpen
  \bibfield  {author} {\bibinfo {author} {\bibfnamefont {J.}~\bibnamefont
  {Yosinski}}, \bibinfo {author} {\bibfnamefont {J.}~\bibnamefont {Clune}},
  \bibinfo {author} {\bibfnamefont {Y.}~\bibnamefont {Bengio}},\ and\ \bibinfo
  {author} {\bibfnamefont {H.}~\bibnamefont {Lipson}},\ }\bibfield  {title}
  {\bibinfo {title} {How transferable are features in deep neural networks?},\
  }in\ \href@noop {} {\emph {\bibinfo {booktitle} {Advances in Neural
  Information Processing Systems (NeurIPS)}}},\ Vol.~\bibinfo {volume} {27}\
  (\bibinfo {year} {2014})\ pp.\ \bibinfo {pages} {3320--3328},\ \Eprint
  {https://arxiv.org/abs/1411.1792} {arXiv:1411.1792} \BibitemShut {NoStop}%
\bibitem [{\citenamefont {Raghu}\ \emph {et~al.}(2017)\citenamefont {Raghu},
  \citenamefont {Gilmer}, \citenamefont {Yosinski},\ and\ \citenamefont
  {Sohl-Dickstein}}]{raghu2017svcca}%
  \BibitemOpen
  \bibfield  {author} {\bibinfo {author} {\bibfnamefont {M.}~\bibnamefont
  {Raghu}}, \bibinfo {author} {\bibfnamefont {J.}~\bibnamefont {Gilmer}},
  \bibinfo {author} {\bibfnamefont {J.}~\bibnamefont {Yosinski}},\ and\
  \bibinfo {author} {\bibfnamefont {J.}~\bibnamefont {Sohl-Dickstein}},\
  }\bibfield  {title} {\bibinfo {title} {{SVCCA}: Singular vector canonical
  correlation analysis for deep learning dynamics and interpretability},\ }in\
  \href@noop {} {\emph {\bibinfo {booktitle} {Advances in Neural Information
  Processing Systems (NeurIPS)}}},\ Vol.~\bibinfo {volume} {30}\ (\bibinfo
  {year} {2017})\ pp.\ \bibinfo {pages} {6076--6085},\ \Eprint
  {https://arxiv.org/abs/1706.05806} {arXiv:1706.05806} \BibitemShut {NoStop}%
\bibitem [{\citenamefont {Saxe}\ \emph {et~al.}(2014)\citenamefont {Saxe},
  \citenamefont {McClelland},\ and\ \citenamefont {Ganguli}}]{saxe2014exact}%
  \BibitemOpen
  \bibfield  {author} {\bibinfo {author} {\bibfnamefont {A.~M.}\ \bibnamefont
  {Saxe}}, \bibinfo {author} {\bibfnamefont {J.~L.}\ \bibnamefont
  {McClelland}},\ and\ \bibinfo {author} {\bibfnamefont {S.}~\bibnamefont
  {Ganguli}},\ }\bibfield  {title} {\bibinfo {title} {Exact solutions to the
  nonlinear dynamics of learning in deep linear neural networks},\ }in\
  \href@noop {} {\emph {\bibinfo {booktitle} {International Conference on
  Learning Representations (ICLR)}}}\ (\bibinfo {year} {2014})\ \Eprint
  {https://arxiv.org/abs/1312.6120} {arXiv:1312.6120} \BibitemShut {NoStop}%
\bibitem [{\citenamefont {Jacot}\ \emph {et~al.}(2018)\citenamefont {Jacot},
  \citenamefont {Gabriel},\ and\ \citenamefont {Hongler}}]{jacot2018neural}%
  \BibitemOpen
  \bibfield  {author} {\bibinfo {author} {\bibfnamefont {A.}~\bibnamefont
  {Jacot}}, \bibinfo {author} {\bibfnamefont {F.}~\bibnamefont {Gabriel}},\
  and\ \bibinfo {author} {\bibfnamefont {C.}~\bibnamefont {Hongler}},\
  }\bibfield  {title} {\bibinfo {title} {Neural tangent kernel: Convergence and
  generalization in neural networks},\ }in\ \href@noop {} {\emph {\bibinfo
  {booktitle} {Advances in Neural Information Processing Systems (NeurIPS)}}},\
  Vol.~\bibinfo {volume} {31}\ (\bibinfo {year} {2018})\ pp.\ \bibinfo {pages}
  {8571--8580},\ \Eprint {https://arxiv.org/abs/1806.07572} {arXiv:1806.07572}
  \BibitemShut {NoStop}%
\bibitem [{\citenamefont {Kaplan}\ \emph {et~al.}(2020)\citenamefont {Kaplan},
  \citenamefont {McCandlish}, \citenamefont {Henighan}, \citenamefont {Brown},
  \citenamefont {Chess}, \citenamefont {Child}, \citenamefont {Gray},
  \citenamefont {Radford}, \citenamefont {Wu},\ and\ \citenamefont
  {Amodei}}]{kaplan2020scaling}%
  \BibitemOpen
  \bibfield  {author} {\bibinfo {author} {\bibfnamefont {J.}~\bibnamefont
  {Kaplan}}, \bibinfo {author} {\bibfnamefont {S.}~\bibnamefont {McCandlish}},
  \bibinfo {author} {\bibfnamefont {T.}~\bibnamefont {Henighan}}, \bibinfo
  {author} {\bibfnamefont {T.~B.}\ \bibnamefont {Brown}}, \bibinfo {author}
  {\bibfnamefont {B.}~\bibnamefont {Chess}}, \bibinfo {author} {\bibfnamefont
  {R.}~\bibnamefont {Child}}, \bibinfo {author} {\bibfnamefont
  {S.}~\bibnamefont {Gray}}, \bibinfo {author} {\bibfnamefont {A.}~\bibnamefont
  {Radford}}, \bibinfo {author} {\bibfnamefont {J.}~\bibnamefont {Wu}},\ and\
  \bibinfo {author} {\bibfnamefont {D.}~\bibnamefont {Amodei}},\ }\href@noop {}
  {\bibinfo {title} {Scaling laws for neural language models}} (\bibinfo {year}
  {2020}),\ \Eprint {https://arxiv.org/abs/2001.08361} {arXiv:2001.08361}
  \BibitemShut {NoStop}%
\bibitem [{\citenamefont {Alt{\i}nta{\c{s}}}\ \emph {et~al.}(2025)\citenamefont
  {Alt{\i}nta{\c{s}}}, \citenamefont {Kwok}, \citenamefont {Raffel},\ and\
  \citenamefont {Rolnick}}]{altintas2025butterfly}%
  \BibitemOpen
  \bibfield  {author} {\bibinfo {author} {\bibfnamefont {G.~S.}\ \bibnamefont
  {Alt{\i}nta{\c{s}}}}, \bibinfo {author} {\bibfnamefont {D.}~\bibnamefont
  {Kwok}}, \bibinfo {author} {\bibfnamefont {C.}~\bibnamefont {Raffel}},\ and\
  \bibinfo {author} {\bibfnamefont {D.}~\bibnamefont {Rolnick}},\ }\bibfield
  {title} {\bibinfo {title} {The butterfly effect: Neural network training
  trajectories are highly sensitive to initial conditions},\ }in\ \href@noop {}
  {\emph {\bibinfo {booktitle} {Proceedings of the 42nd International
  Conference on Machine Learning (ICML)}}},\ \bibinfo {series} {Proceedings of
  Machine Learning Research}, Vol.\ \bibinfo {volume} {267}\ (\bibinfo {year}
  {2025})\ pp.\ \bibinfo {pages} {1314--1342},\ \Eprint
  {https://arxiv.org/abs/2506.13234} {arXiv:2506.13234} \BibitemShut {NoStop}%
\bibitem [{\citenamefont {Lin}\ and\ \citenamefont
  {Wang}(2026)}]{lin2026mapping}%
  \BibitemOpen
  \bibfield  {author} {\bibinfo {author} {\bibfnamefont {Y.}~\bibnamefont
  {Lin}}\ and\ \bibinfo {author} {\bibfnamefont {Y.}~\bibnamefont {Wang}},\
  }\href@noop {} {\bibinfo {title} {Mapping the emergence of
  regularization-driven dynamics in grokking}} (\bibinfo {year} {2026}),\
  \Eprint {https://arxiv.org/abs/2608.25813} {arXiv:2608.25813} \BibitemShut
  {NoStop}%
\bibitem [{\citenamefont {Power}\ \emph {et~al.}(2022)\citenamefont {Power},
  \citenamefont {Burda}, \citenamefont {Edwards}, \citenamefont {Babuschkin},\
  and\ \citenamefont {Misra}}]{power2022grokking}%
  \BibitemOpen
  \bibfield  {author} {\bibinfo {author} {\bibfnamefont {A.}~\bibnamefont
  {Power}}, \bibinfo {author} {\bibfnamefont {Y.}~\bibnamefont {Burda}},
  \bibinfo {author} {\bibfnamefont {H.}~\bibnamefont {Edwards}}, \bibinfo
  {author} {\bibfnamefont {I.}~\bibnamefont {Babuschkin}},\ and\ \bibinfo
  {author} {\bibfnamefont {V.}~\bibnamefont {Misra}},\ }\href@noop {} {\bibinfo
  {title} {Grokking: Generalization beyond overfitting on small algorithmic
  datasets}} (\bibinfo {year} {2022}),\ \Eprint
  {https://arxiv.org/abs/2201.02177} {arXiv:2201.02177} \BibitemShut {NoStop}%
\bibitem [{\citenamefont {Liu}\ \emph {et~al.}(2023)\citenamefont {Liu},
  \citenamefont {Michaud},\ and\ \citenamefont {Tegmark}}]{liu2022omnigrok}%
  \BibitemOpen
  \bibfield  {author} {\bibinfo {author} {\bibfnamefont {Z.}~\bibnamefont
  {Liu}}, \bibinfo {author} {\bibfnamefont {E.~J.}\ \bibnamefont {Michaud}},\
  and\ \bibinfo {author} {\bibfnamefont {M.}~\bibnamefont {Tegmark}},\
  }\bibfield  {title} {\bibinfo {title} {Omnigrok: Grokking beyond algorithmic
  data},\ }in\ \href@noop {} {\emph {\bibinfo {booktitle} {International
  Conference on Learning Representations (ICLR)}}}\ (\bibinfo {year} {2023})\
  \Eprint {https://arxiv.org/abs/2210.01117} {arXiv:2210.01117} \BibitemShut
  {NoStop}%
\bibitem [{\citenamefont {Nanda}\ \emph {et~al.}(2023)\citenamefont {Nanda},
  \citenamefont {Chan}, \citenamefont {Lieberum}, \citenamefont {Smith},\ and\
  \citenamefont {Steinhardt}}]{nanda2023progress}%
  \BibitemOpen
  \bibfield  {author} {\bibinfo {author} {\bibfnamefont {N.}~\bibnamefont
  {Nanda}}, \bibinfo {author} {\bibfnamefont {L.}~\bibnamefont {Chan}},
  \bibinfo {author} {\bibfnamefont {T.}~\bibnamefont {Lieberum}}, \bibinfo
  {author} {\bibfnamefont {J.}~\bibnamefont {Smith}},\ and\ \bibinfo {author}
  {\bibfnamefont {J.}~\bibnamefont {Steinhardt}},\ }\bibfield  {title}
  {\bibinfo {title} {Progress measures for grokking via mechanistic
  interpretability},\ }in\ \href@noop {} {\emph {\bibinfo {booktitle}
  {International Conference on Learning Representations (ICLR)}}}\ (\bibinfo
  {year} {2023})\ \Eprint {https://arxiv.org/abs/2301.05217} {arXiv:2301.05217}
  \BibitemShut {NoStop}%
\bibitem [{\citenamefont {Yunis}\ \emph {et~al.}(2024)\citenamefont {Yunis},
  \citenamefont {Patel}, \citenamefont {Wheeler}, \citenamefont {Savarese},
  \citenamefont {Vardi}, \citenamefont {Livescu}, \citenamefont {Maire},\ and\
  \citenamefont {Walter}}]{yunis2024spectral}%
  \BibitemOpen
  \bibfield  {author} {\bibinfo {author} {\bibfnamefont {D.}~\bibnamefont
  {Yunis}}, \bibinfo {author} {\bibfnamefont {K.~K.}\ \bibnamefont {Patel}},
  \bibinfo {author} {\bibfnamefont {S.}~\bibnamefont {Wheeler}}, \bibinfo
  {author} {\bibfnamefont {P.}~\bibnamefont {Savarese}}, \bibinfo {author}
  {\bibfnamefont {G.}~\bibnamefont {Vardi}}, \bibinfo {author} {\bibfnamefont
  {K.}~\bibnamefont {Livescu}}, \bibinfo {author} {\bibfnamefont
  {M.}~\bibnamefont {Maire}},\ and\ \bibinfo {author} {\bibfnamefont {M.~R.}\
  \bibnamefont {Walter}},\ }\href@noop {} {\bibinfo {title} {Approaching deep
  learning through the spectral dynamics of weights}} (\bibinfo {year}
  {2024}),\ \Eprint {https://arxiv.org/abs/2408.11804} {arXiv:2408.11804}
  \BibitemShut {NoStop}%
\bibitem [{\citenamefont {Clauw}\ \emph {et~al.}(2024)\citenamefont {Clauw},
  \citenamefont {Stramaglia},\ and\ \citenamefont
  {Marinazzo}}]{clauw2024information}%
  \BibitemOpen
  \bibfield  {author} {\bibinfo {author} {\bibfnamefont {K.}~\bibnamefont
  {Clauw}}, \bibinfo {author} {\bibfnamefont {S.}~\bibnamefont {Stramaglia}},\
  and\ \bibinfo {author} {\bibfnamefont {D.}~\bibnamefont {Marinazzo}},\
  }\href@noop {} {\bibinfo {title} {Information-theoretic progress measures
  reveal grokking is an emergent phase transition}} (\bibinfo {year} {2024}),\
  \bibinfo {note} {{ICML} 2024 Workshop on Mechanistic Interpretability},\
  \Eprint {https://arxiv.org/abs/2408.08944} {arXiv:2408.08944} \BibitemShut
  {NoStop}%
\bibitem [{\citenamefont {Marrow}\ \emph {et~al.}(2020)\citenamefont {Marrow},
  \citenamefont {Michaud},\ and\ \citenamefont {Hoel}}]{mattsson2020examining}%
  \BibitemOpen
  \bibfield  {author} {\bibinfo {author} {\bibfnamefont {S.}~\bibnamefont
  {Marrow}}, \bibinfo {author} {\bibfnamefont {E.~J.}\ \bibnamefont
  {Michaud}},\ and\ \bibinfo {author} {\bibfnamefont {E.}~\bibnamefont
  {Hoel}},\ }\bibfield  {title} {\bibinfo {title} {Examining the causal
  structures of deep neural networks using information theory},\ }\href
  {https://doi.org/10.3390/e22121429} {\bibfield  {journal} {\bibinfo
  {journal} {Entropy}\ }\textbf {\bibinfo {volume} {22}},\ \bibinfo {pages}
  {1429} (\bibinfo {year} {2020})}\BibitemShut {NoStop}%
\bibitem [{\citenamefont {Moldovan}\ \emph {et~al.}(2023)\citenamefont
  {Moldovan}, \citenamefont {Ca{\c{t}}aron},\ and\ \citenamefont
  {Andonie}}]{moldovan2023information}%
  \BibitemOpen
  \bibfield  {author} {\bibinfo {author} {\bibfnamefont {A.}~\bibnamefont
  {Moldovan}}, \bibinfo {author} {\bibfnamefont {A.}~\bibnamefont
  {Ca{\c{t}}aron}},\ and\ \bibinfo {author} {\bibfnamefont {R.}~\bibnamefont
  {Andonie}},\ }\bibfield  {title} {\bibinfo {title} {Information plane
  analysis visualization in deep learning via transfer entropy},\ }in\ \href
  {https://doi.org/10.1109/IV60283.2023.00055} {\emph {\bibinfo {booktitle}
  {27th International Conference Information Visualisation (IV)}}}\ (\bibinfo
  {year} {2023})\ pp.\ \bibinfo {pages} {278--285},\ \Eprint
  {https://arxiv.org/abs/2404.01364} {arXiv:2404.01364} \BibitemShut {NoStop}%
\bibitem [{\citenamefont {Wang}\ \emph {et~al.}(2024)\citenamefont {Wang},
  \citenamefont {Yue}, \citenamefont {Su},\ and\ \citenamefont
  {Sun}}]{wang2024grokked}%
  \BibitemOpen
  \bibfield  {author} {\bibinfo {author} {\bibfnamefont {B.}~\bibnamefont
  {Wang}}, \bibinfo {author} {\bibfnamefont {X.}~\bibnamefont {Yue}}, \bibinfo
  {author} {\bibfnamefont {Y.}~\bibnamefont {Su}},\ and\ \bibinfo {author}
  {\bibfnamefont {H.}~\bibnamefont {Sun}},\ }\bibfield  {title} {\bibinfo
  {title} {Grokking of implicit reasoning in transformers: A mechanistic
  journey to the edge of generalization},\ }in\ \href@noop {} {\emph {\bibinfo
  {booktitle} {Advances in Neural Information Processing Systems (NeurIPS)}}},\
  Vol.~\bibinfo {volume} {37}\ (\bibinfo {year} {2024})\ pp.\ \bibinfo {pages}
  {95238--95265},\ \Eprint {https://arxiv.org/abs/2405.15071}
  {arXiv:2405.15071} \BibitemShut {NoStop}%
\bibitem [{\citenamefont {Tao}\ \emph {et~al.}(2026)\citenamefont {Tao},
  \citenamefont {Cheng},\ and\ \citenamefont {Xie}}]{tao2026persistent}%
  \BibitemOpen
  \bibfield  {author} {\bibinfo {author} {\bibfnamefont {S.}~\bibnamefont
  {Tao}}, \bibinfo {author} {\bibfnamefont {Y.}~\bibnamefont {Cheng}},\ and\
  \bibinfo {author} {\bibfnamefont {H.}~\bibnamefont {Xie}},\ }\href@noop {}
  {\bibinfo {title} {Persistent depth ordering amid shifting block-bypass
  responses in language model pretraining}} (\bibinfo {year} {2026}),\ \Eprint
  {https://arxiv.org/abs/2610.01165} {arXiv:2610.01165} \BibitemShut {NoStop}%
\bibitem [{\citenamefont {Ji}\ \emph {et~al.}(2026)\citenamefont {Ji},
  \citenamefont {Wang},\ and\ \citenamefont {Sun}}]{ji2026bridging}%
  \BibitemOpen
  \bibfield  {author} {\bibinfo {author} {\bibfnamefont {E.}~\bibnamefont
  {Ji}}, \bibinfo {author} {\bibfnamefont {W.}~\bibnamefont {Wang}},\ and\
  \bibinfo {author} {\bibfnamefont {Y.}~\bibnamefont {Sun}},\ }\href@noop {}
  {\bibinfo {title} {Bridging singular learning theory and mechanistic
  interpretability at the billion-parameter scale}} (\bibinfo {year} {2026}),\
  \Eprint {https://arxiv.org/abs/2611.00640} {arXiv:2611.00640} \BibitemShut
  {NoStop}%
\bibitem [{\citenamefont {Lan}\ \emph {et~al.}(2019)\citenamefont {Lan},
  \citenamefont {Liu}, \citenamefont {Zhou},\ and\ \citenamefont
  {Yosinski}}]{lan2019lca}%
  \BibitemOpen
  \bibfield  {author} {\bibinfo {author} {\bibfnamefont {J.}~\bibnamefont
  {Lan}}, \bibinfo {author} {\bibfnamefont {R.}~\bibnamefont {Liu}}, \bibinfo
  {author} {\bibfnamefont {H.}~\bibnamefont {Zhou}},\ and\ \bibinfo {author}
  {\bibfnamefont {J.}~\bibnamefont {Yosinski}},\ }\bibfield  {title} {\bibinfo
  {title} {{LCA}: Loss change allocation for neural network training},\ }in\
  \href@noop {} {\emph {\bibinfo {booktitle} {Advances in Neural Information
  Processing Systems (NeurIPS)}}},\ Vol.~\bibinfo {volume} {32}\ (\bibinfo
  {year} {2019})\ pp.\ \bibinfo {pages} {3619--3629},\ \Eprint
  {https://arxiv.org/abs/1909.01440} {arXiv:1909.01440} \BibitemShut {NoStop}%
\bibitem [{\citenamefont {Allen-Zhu}\ and\ \citenamefont
  {Li}(2023)}]{allenzhu2023backward}%
  \BibitemOpen
  \bibfield  {author} {\bibinfo {author} {\bibfnamefont {Z.}~\bibnamefont
  {Allen-Zhu}}\ and\ \bibinfo {author} {\bibfnamefont {Y.}~\bibnamefont {Li}},\
  }\bibfield  {title} {\bibinfo {title} {Backward feature correction: How deep
  learning performs deep (hierarchical) learning},\ }in\ \href@noop {} {\emph
  {\bibinfo {booktitle} {Proceedings of the 36th Conference on Learning Theory
  (COLT)}}},\ \bibinfo {series} {Proceedings of Machine Learning Research},
  Vol.\ \bibinfo {volume} {195}\ (\bibinfo {year} {2023})\ p.\ \bibinfo {pages}
  {4598},\ \bibinfo {note} {extended abstract},\ \Eprint
  {https://arxiv.org/abs/2001.04413} {arXiv:2001.04413} \BibitemShut {NoStop}%
\bibitem [{\citenamefont {Loshchilov}\ and\ \citenamefont
  {Hutter}(2019)}]{loshchilov2019decoupled}%
  \BibitemOpen
  \bibfield  {author} {\bibinfo {author} {\bibfnamefont {I.}~\bibnamefont
  {Loshchilov}}\ and\ \bibinfo {author} {\bibfnamefont {F.}~\bibnamefont
  {Hutter}},\ }\bibfield  {title} {\bibinfo {title} {Decoupled weight decay
  regularization},\ }in\ \href@noop {} {\emph {\bibinfo {booktitle}
  {International Conference on Learning Representations (ICLR)}}}\ (\bibinfo
  {year} {2019})\ \Eprint {https://arxiv.org/abs/1711.05101} {arXiv:1711.05101}
  \BibitemShut {NoStop}%
\bibitem [{\citenamefont {Welling}\ and\ \citenamefont
  {Teh}(2011)}]{welling2011bayesian}%
  \BibitemOpen
  \bibfield  {author} {\bibinfo {author} {\bibfnamefont {M.}~\bibnamefont
  {Welling}}\ and\ \bibinfo {author} {\bibfnamefont {Y.~W.}\ \bibnamefont
  {Teh}},\ }\bibfield  {title} {\bibinfo {title} {Bayesian learning via
  stochastic gradient {L}angevin dynamics},\ }in\ \href@noop {} {\emph
  {\bibinfo {booktitle} {Proceedings of the 28th International Conference on
  Machine Learning (ICML)}}}\ (\bibinfo {year} {2011})\ pp.\ \bibinfo {pages}
  {681--688}\BibitemShut {NoStop}%
\bibitem [{\citenamefont {Raginsky}\ \emph {et~al.}(2017)\citenamefont
  {Raginsky}, \citenamefont {Rakhlin},\ and\ \citenamefont
  {Telgarsky}}]{raginsky2017nonconvex}%
  \BibitemOpen
  \bibfield  {author} {\bibinfo {author} {\bibfnamefont {M.}~\bibnamefont
  {Raginsky}}, \bibinfo {author} {\bibfnamefont {A.}~\bibnamefont {Rakhlin}},\
  and\ \bibinfo {author} {\bibfnamefont {M.}~\bibnamefont {Telgarsky}},\
  }\bibfield  {title} {\bibinfo {title} {Non-convex learning via stochastic
  gradient {Langevin} dynamics: {A} nonasymptotic analysis},\ }in\ \href@noop
  {} {\emph {\bibinfo {booktitle} {Proceedings of the 2017 Conference on
  Learning Theory (COLT)}}},\ \bibinfo {series} {Proceedings of Machine
  Learning Research}, Vol.~\bibinfo {volume} {65}\ (\bibinfo {year} {2017})\
  pp.\ \bibinfo {pages} {1674--1703},\ \Eprint
  {https://arxiv.org/abs/1702.03849} {arXiv:1702.03849} \BibitemShut {NoStop}%
\bibitem [{\citenamefont {Aubry}\ \emph {et~al.}(2025)\citenamefont {Aubry},
  \citenamefont {Meng}, \citenamefont {Sugolov},\ and\ \citenamefont
  {Papyan}}]{aubry2025coupling}%
  \BibitemOpen
  \bibfield  {author} {\bibinfo {author} {\bibfnamefont {M.}~\bibnamefont
  {Aubry}}, \bibinfo {author} {\bibfnamefont {H.}~\bibnamefont {Meng}},
  \bibinfo {author} {\bibfnamefont {A.}~\bibnamefont {Sugolov}},\ and\ \bibinfo
  {author} {\bibfnamefont {V.}~\bibnamefont {Papyan}},\ }\bibfield  {title}
  {\bibinfo {title} {Transformer block coupling and its correlation with
  generalization in {LLMs}},\ }in\ \href@noop {} {\emph {\bibinfo {booktitle}
  {International Conference on Learning Representations (ICLR)}}}\ (\bibinfo
  {year} {2025})\ \Eprint {https://arxiv.org/abs/2407.07810} {arXiv:2407.07810}
  \BibitemShut {NoStop}%
\bibitem [{\citenamefont {Li}\ and\ \citenamefont
  {Papyan}(2023)}]{li2023residual}%
  \BibitemOpen
  \bibfield  {author} {\bibinfo {author} {\bibfnamefont {J.}~\bibnamefont
  {Li}}\ and\ \bibinfo {author} {\bibfnamefont {V.}~\bibnamefont {Papyan}},\
  }\bibfield  {title} {\bibinfo {title} {Residual alignment: Uncovering the
  mechanisms of residual networks},\ }in\ \href@noop {} {\emph {\bibinfo
  {booktitle} {Advances in Neural Information Processing Systems (NeurIPS)}}},\
  Vol.~\bibinfo {volume} {36}\ (\bibinfo {year} {2023})\ pp.\ \bibinfo {pages}
  {57660--57712},\ \Eprint {https://arxiv.org/abs/2401.09018}
  {arXiv:2401.09018} \BibitemShut {NoStop}%
\bibitem [{\citenamefont {Lee}\ \emph {et~al.}(2025)\citenamefont {Lee},
  \citenamefont {Lee},\ and\ \citenamefont {Yi}}]{lee2025bulk}%
  \BibitemOpen
  \bibfield  {author} {\bibinfo {author} {\bibfnamefont {D.}~\bibnamefont
  {Lee}}, \bibinfo {author} {\bibfnamefont {H.-S.}\ \bibnamefont {Lee}},\ and\
  \bibinfo {author} {\bibfnamefont {J.}~\bibnamefont {Yi}},\ }\href@noop {}
  {\bibinfo {title} {Bulk-boundary decomposition of neural networks}} (\bibinfo
  {year} {2025}),\ \Eprint {https://arxiv.org/abs/2511.02003}
  {arXiv:2511.02003} \BibitemShut {NoStop}%
\bibitem [{\citenamefont {Bolshim}\ and\ \citenamefont
  {Kugaevskikh}(2026)}]{bolshim2026interlayer}%
  \BibitemOpen
  \bibfield  {author} {\bibinfo {author} {\bibfnamefont {M.}~\bibnamefont
  {Bolshim}}\ and\ \bibinfo {author} {\bibfnamefont {A.}~\bibnamefont
  {Kugaevskikh}},\ }\href@noop {} {\bibinfo {title} {Inter-layer {H}essian
  analysis of neural networks with {DAG} architectures}} (\bibinfo {year}
  {2026}),\ \Eprint {https://arxiv.org/abs/2604.11639} {arXiv:2604.11639}
  \BibitemShut {NoStop}%
\bibitem [{\citenamefont {de~Jong}\ \emph {et~al.}(2025)\citenamefont
  {de~Jong}, \citenamefont {Meigel},\ and\ \citenamefont
  {Rulands}}]{dejong2025emergent}%
  \BibitemOpen
  \bibfield  {author} {\bibinfo {author} {\bibfnamefont {P.}~\bibnamefont
  {de~Jong}}, \bibinfo {author} {\bibfnamefont {F.~J.}\ \bibnamefont
  {Meigel}},\ and\ \bibinfo {author} {\bibfnamefont {S.}~\bibnamefont
  {Rulands}},\ }\href@noop {} {\bibinfo {title} {Emergent weight morphologies
  in deep neural networks}} (\bibinfo {year} {2025}),\ \Eprint
  {https://arxiv.org/abs/2501.05550} {arXiv:2501.05550} \BibitemShut {NoStop}%
\bibitem [{\citenamefont {Ji}\ and\ \citenamefont
  {Telgarsky}(2019)}]{ji2018aligns}%
  \BibitemOpen
  \bibfield  {author} {\bibinfo {author} {\bibfnamefont {Z.}~\bibnamefont
  {Ji}}\ and\ \bibinfo {author} {\bibfnamefont {M.}~\bibnamefont {Telgarsky}},\
  }\bibfield  {title} {\bibinfo {title} {Gradient descent aligns the layers of
  deep linear networks},\ }in\ \href@noop {} {\emph {\bibinfo {booktitle}
  {International Conference on Learning Representations (ICLR)}}}\ (\bibinfo
  {year} {2019})\ \Eprint {https://arxiv.org/abs/1810.02032} {arXiv:1810.02032}
  \BibitemShut {NoStop}%
\bibitem [{\citenamefont {Yunis}(2026)}]{yunis2026thesis}%
  \BibitemOpen
  \bibfield  {author} {\bibinfo {author} {\bibfnamefont {D.~A.}\ \bibnamefont
  {Yunis}},\ }\emph {\bibinfo {title} {From the Loss Landscape to Diverse
  Feature Learning in Neural Networks}},\ \href@noop {} {Ph.D. thesis},\
  \bibinfo  {school} {Toyota Technological Institute at Chicago} (\bibinfo
  {year} {2026}),\ \Eprint {https://arxiv.org/abs/2608.28948}
  {arXiv:2608.28948} \BibitemShut {NoStop}%
\bibitem [{\citenamefont {Jeong}\ \emph {et~al.}(2026)\citenamefont {Jeong},
  \citenamefont {Paeng},\ and\ \citenamefont {Suh}}]{jeong2026geometric}%
  \BibitemOpen
  \bibfield  {author} {\bibinfo {author} {\bibfnamefont {K.}~\bibnamefont
  {Jeong}}, \bibinfo {author} {\bibfnamefont {W.-G.}\ \bibnamefont {Paeng}},\
  and\ \bibinfo {author} {\bibfnamefont {H.}~\bibnamefont {Suh}},\ }\bibfield
  {title} {\bibinfo {title} {Why geometric continuity emerges in deep neural
  networks: Residual connections and rotational symmetry breaking},\ }in\
  \href@noop {} {\emph {\bibinfo {booktitle} {Advances in Neural Information
  Processing Systems (NeurIPS)}}}\ (\bibinfo {year} {2026})\ \Eprint
  {https://arxiv.org/abs/2605.04971} {arXiv:2605.04971} \BibitemShut {NoStop}%
\bibitem [{\citenamefont {Bray}(2026)}]{bray2026common}%
  \BibitemOpen
  \bibfield  {author} {\bibinfo {author} {\bibfnamefont {D.}~\bibnamefont
  {Bray}, \bibfnamefont {Jr.}},\ }\bibfield  {title} {\bibinfo {title} {Common
  origins, divergent destinations: The development of cross-layer alignment
  under {GELU} and {SwiGLU}},\ }in\ \href
  {https://openreview.net/forum?id=8MDrWfFj6H} {\emph {\bibinfo {booktitle}
  {International Conference on Machine Learning (ICML) Workshop on High
  Dimensional Learning Dynamics: the Science of Scaling}}}\ (\bibinfo {year}
  {2026})\BibitemShut {NoStop}%
\bibitem [{\citenamefont {Arora}\ \emph {et~al.}(2018)\citenamefont {Arora},
  \citenamefont {Cohen},\ and\ \citenamefont {Hazan}}]{arora2018optimization}%
  \BibitemOpen
  \bibfield  {author} {\bibinfo {author} {\bibfnamefont {S.}~\bibnamefont
  {Arora}}, \bibinfo {author} {\bibfnamefont {N.}~\bibnamefont {Cohen}},\ and\
  \bibinfo {author} {\bibfnamefont {E.}~\bibnamefont {Hazan}},\ }\bibfield
  {title} {\bibinfo {title} {On the optimization of deep networks: Implicit
  acceleration by overparameterization},\ }in\ \href@noop {} {\emph {\bibinfo
  {booktitle} {Proceedings of the 35th International Conference on Machine
  Learning (ICML)}}},\ \bibinfo {series} {Proceedings of Machine Learning
  Research}, Vol.~\bibinfo {volume} {80}\ (\bibinfo {year} {2018})\ pp.\
  \bibinfo {pages} {244--253},\ \Eprint {https://arxiv.org/abs/1802.06509}
  {arXiv:1802.06509} \BibitemShut {NoStop}%
\bibitem [{\citenamefont {Du}\ \emph {et~al.}(2018)\citenamefont {Du},
  \citenamefont {Hu},\ and\ \citenamefont {Lee}}]{du2018algorithmic}%
  \BibitemOpen
  \bibfield  {author} {\bibinfo {author} {\bibfnamefont {S.~S.}\ \bibnamefont
  {Du}}, \bibinfo {author} {\bibfnamefont {W.}~\bibnamefont {Hu}},\ and\
  \bibinfo {author} {\bibfnamefont {J.~D.}\ \bibnamefont {Lee}},\ }\bibfield
  {title} {\bibinfo {title} {Algorithmic regularization in learning deep
  homogeneous models: Layers are automatically balanced},\ }in\ \href@noop {}
  {\emph {\bibinfo {booktitle} {Advances in Neural Information Processing
  Systems (NeurIPS)}}},\ Vol.~\bibinfo {volume} {31}\ (\bibinfo {year} {2018})\
  pp.\ \bibinfo {pages} {384--395},\ \Eprint {https://arxiv.org/abs/1806.00900}
  {arXiv:1806.00900} \BibitemShut {NoStop}%
\bibitem [{\citenamefont {Le}\ and\ \citenamefont
  {Jegelka}(2022)}]{le2022training}%
  \BibitemOpen
  \bibfield  {author} {\bibinfo {author} {\bibfnamefont {T.}~\bibnamefont
  {Le}}\ and\ \bibinfo {author} {\bibfnamefont {S.}~\bibnamefont {Jegelka}},\
  }\bibfield  {title} {\bibinfo {title} {Training invariances and the low-rank
  phenomenon: {B}eyond linear networks},\ }in\ \href@noop {} {\emph {\bibinfo
  {booktitle} {International Conference on Learning Representations (ICLR)}}}\
  (\bibinfo {year} {2022})\ \Eprint {https://arxiv.org/abs/2201.11968}
  {arXiv:2201.11968} \BibitemShut {NoStop}%
\bibitem [{\citenamefont {Rawal}\ and\ \citenamefont
  {DeWeese}(2026)}]{rawal2026saddle}%
  \BibitemOpen
  \bibfield  {author} {\bibinfo {author} {\bibfnamefont {D.}~\bibnamefont
  {Rawal}}\ and\ \bibinfo {author} {\bibfnamefont {M.~R.}\ \bibnamefont
  {DeWeese}},\ }\href@noop {} {\bibinfo {title} {A theory of saddle escape in
  deep nonlinear networks}} (\bibinfo {year} {2026}),\ \Eprint
  {https://arxiv.org/abs/2605.01288v4} {arXiv:2605.01288v4} \BibitemShut
  {NoStop}%
\bibitem [{\citenamefont {Nystr{\"o}m}(2026)}]{nystrom2026commutator}%
  \BibitemOpen
  \bibfield  {author} {\bibinfo {author} {\bibfnamefont {K.}~\bibnamefont
  {Nystr{\"o}m}},\ }\href@noop {} {\bibinfo {title} {A commutator framework for
  selective spectral alignment in deep neural networks}} (\bibinfo {year}
  {2026}),\ \Eprint {https://arxiv.org/abs/2608.22910} {arXiv:2608.22910}
  \BibitemShut {NoStop}%
\bibitem [{\citenamefont {Varma}\ \emph {et~al.}(2023)\citenamefont {Varma},
  \citenamefont {Shah}, \citenamefont {Kenton}, \citenamefont {Kram{\'a}r},\
  and\ \citenamefont {Kumar}}]{varma2023explaining}%
  \BibitemOpen
  \bibfield  {author} {\bibinfo {author} {\bibfnamefont {V.}~\bibnamefont
  {Varma}}, \bibinfo {author} {\bibfnamefont {R.}~\bibnamefont {Shah}},
  \bibinfo {author} {\bibfnamefont {Z.}~\bibnamefont {Kenton}}, \bibinfo
  {author} {\bibfnamefont {J.}~\bibnamefont {Kram{\'a}r}},\ and\ \bibinfo
  {author} {\bibfnamefont {R.}~\bibnamefont {Kumar}},\ }\href@noop {} {\bibinfo
  {title} {Explaining grokking through circuit efficiency}} (\bibinfo {year}
  {2023}),\ \Eprint {https://arxiv.org/abs/2309.02390} {arXiv:2309.02390}
  \BibitemShut {NoStop}%
\bibitem [{\citenamefont {Prieto}\ \emph {et~al.}(2025)\citenamefont {Prieto},
  \citenamefont {Barsbey}, \citenamefont {Mediano},\ and\ \citenamefont
  {Birdal}}]{prieto2025grokking}%
  \BibitemOpen
  \bibfield  {author} {\bibinfo {author} {\bibfnamefont {L.}~\bibnamefont
  {Prieto}}, \bibinfo {author} {\bibfnamefont {M.}~\bibnamefont {Barsbey}},
  \bibinfo {author} {\bibfnamefont {P.~A.~M.}\ \bibnamefont {Mediano}},\ and\
  \bibinfo {author} {\bibfnamefont {T.}~\bibnamefont {Birdal}},\ }\bibfield
  {title} {\bibinfo {title} {Grokking at the edge of numerical stability},\
  }in\ \href@noop {} {\emph {\bibinfo {booktitle} {International Conference on
  Learning Representations (ICLR)}}}\ (\bibinfo {year} {2025})\ \Eprint
  {https://arxiv.org/abs/2501.04697} {arXiv:2501.04697} \BibitemShut {NoStop}%
\bibitem [{\citenamefont {Liu}\ \emph {et~al.}(2026)\citenamefont {Liu},
  \citenamefont {Cao}, \citenamefont {Li},\ and\ \citenamefont
  {Zhou}}]{liu2026grokking}%
  \BibitemOpen
  \bibfield  {author} {\bibinfo {author} {\bibfnamefont {H.}~\bibnamefont
  {Liu}}, \bibinfo {author} {\bibfnamefont {J.}~\bibnamefont {Cao}}, \bibinfo
  {author} {\bibfnamefont {Y.}~\bibnamefont {Li}},\ and\ \bibinfo {author}
  {\bibfnamefont {Z.}~\bibnamefont {Zhou}},\ }\bibfield  {title} {\bibinfo
  {title} {Grokking or glitching? {H}ow low-precision drives slingshot loss
  spikes},\ }in\ \href@noop {} {\emph {\bibinfo {booktitle} {Advances in Neural
  Information Processing Systems (NeurIPS)}}}\ (\bibinfo {year} {2026})\
  \Eprint {https://arxiv.org/abs/2605.06152} {arXiv:2605.06152} \BibitemShut
  {NoStop}%
\bibitem [{\citenamefont {Cohen}\ \emph {et~al.}(2021)\citenamefont {Cohen},
  \citenamefont {Kaur}, \citenamefont {Li}, \citenamefont {Kolter},\ and\
  \citenamefont {Talwalkar}}]{cohen2021edge}%
  \BibitemOpen
  \bibfield  {author} {\bibinfo {author} {\bibfnamefont {J.~M.}\ \bibnamefont
  {Cohen}}, \bibinfo {author} {\bibfnamefont {S.}~\bibnamefont {Kaur}},
  \bibinfo {author} {\bibfnamefont {Y.}~\bibnamefont {Li}}, \bibinfo {author}
  {\bibfnamefont {J.~Z.}\ \bibnamefont {Kolter}},\ and\ \bibinfo {author}
  {\bibfnamefont {A.}~\bibnamefont {Talwalkar}},\ }\bibfield  {title} {\bibinfo
  {title} {Gradient descent on neural networks typically occurs at the edge of
  stability},\ }in\ \href@noop {} {\emph {\bibinfo {booktitle} {International
  Conference on Learning Representations (ICLR)}}}\ (\bibinfo {year} {2021})\
  \Eprint {https://arxiv.org/abs/2103.00065} {arXiv:2103.00065} \BibitemShut
  {NoStop}%
\bibitem [{\citenamefont {Lewkowycz}\ \emph {et~al.}(2020)\citenamefont
  {Lewkowycz}, \citenamefont {Bahri}, \citenamefont {Dyer}, \citenamefont
  {Sohl-Dickstein},\ and\ \citenamefont {Gur-Ari}}]{lewkowycz2020catapult}%
  \BibitemOpen
  \bibfield  {author} {\bibinfo {author} {\bibfnamefont {A.}~\bibnamefont
  {Lewkowycz}}, \bibinfo {author} {\bibfnamefont {Y.}~\bibnamefont {Bahri}},
  \bibinfo {author} {\bibfnamefont {E.}~\bibnamefont {Dyer}}, \bibinfo {author}
  {\bibfnamefont {J.}~\bibnamefont {Sohl-Dickstein}},\ and\ \bibinfo {author}
  {\bibfnamefont {G.}~\bibnamefont {Gur-Ari}},\ }\href@noop {} {\bibinfo
  {title} {The large learning rate phase of deep learning: {T}he catapult
  mechanism}} (\bibinfo {year} {2020}),\ \Eprint
  {https://arxiv.org/abs/2003.02218} {arXiv:2003.02218} \BibitemShut {NoStop}%
\bibitem [{\citenamefont {Martens}(2020)}]{martens2020new}%
  \BibitemOpen
  \bibfield  {author} {\bibinfo {author} {\bibfnamefont {J.}~\bibnamefont
  {Martens}},\ }\bibfield  {title} {\bibinfo {title} {New insights and
  perspectives on the natural gradient method},\ }\href@noop {} {\bibfield
  {journal} {\bibinfo  {journal} {J. Mach. Learn. Res.}\ }\textbf {\bibinfo
  {volume} {21}},\ \bibinfo {pages} {146} (\bibinfo {year} {2020})}\BibitemShut
  {NoStop}%
\bibitem [{\citenamefont {Martens}\ and\ \citenamefont
  {Grosse}(2015)}]{martens2015kfac}%
  \BibitemOpen
  \bibfield  {author} {\bibinfo {author} {\bibfnamefont {J.}~\bibnamefont
  {Martens}}\ and\ \bibinfo {author} {\bibfnamefont {R.}~\bibnamefont
  {Grosse}},\ }\bibfield  {title} {\bibinfo {title} {Optimizing neural networks
  with {K}ronecker-factored approximate curvature},\ }in\ \href@noop {} {\emph
  {\bibinfo {booktitle} {Proceedings of the 32nd International Conference on
  Machine Learning (ICML)}}},\ \bibinfo {series} {Proceedings of Machine
  Learning Research}, Vol.~\bibinfo {volume} {37}\ (\bibinfo {year} {2015})\
  pp.\ \bibinfo {pages} {2408--2417},\ \Eprint
  {https://arxiv.org/abs/1503.05671} {arXiv:1503.05671} \BibitemShut {NoStop}%
\bibitem [{\citenamefont {Lee}(2026)}]{lee2026repository}%
  \BibitemOpen
  \bibfield  {author} {\bibinfo {author} {\bibfnamefont {J.}~\bibnamefont
  {Lee}},\ }\href@noop {} {\bibinfo {title} {Data and scripts}},\ \bibinfo
  {howpublished}
  {\url{https://github.com/anffl0101/Pub_Local_interaction_in_NN}} (\bibinfo
  {year} {2026})\BibitemShut {NoStop}%
\bibitem [{\citenamefont {Cugliandolo}\ and\ \citenamefont
  {Kurchan}(1993)}]{Cugliandolo_1993}%
  \BibitemOpen
  \bibfield  {author} {\bibinfo {author} {\bibfnamefont {L.~F.}\ \bibnamefont
  {Cugliandolo}}\ and\ \bibinfo {author} {\bibfnamefont {J.}~\bibnamefont
  {Kurchan}},\ }\bibfield  {title} {\bibinfo {title} {Analytical solution of
  the off-equilibrium dynamics of a long-range spin-glass model},\ }\href
  {https://doi.org/10.1103/physrevlett.71.173} {\bibfield  {journal} {\bibinfo
  {journal} {Phys. Rev. Lett.}\ }\textbf {\bibinfo {volume} {71}},\ \bibinfo
  {pages} {173} (\bibinfo {year} {1993})}\BibitemShut {NoStop}%
\bibitem [{\citenamefont {Benjamini}\ and\ \citenamefont
  {Hochberg}(1995)}]{benjamini1995controlling}%
  \BibitemOpen
  \bibfield  {author} {\bibinfo {author} {\bibfnamefont {Y.}~\bibnamefont
  {Benjamini}}\ and\ \bibinfo {author} {\bibfnamefont {Y.}~\bibnamefont
  {Hochberg}},\ }\bibfield  {title} {\bibinfo {title} {Controlling the false
  discovery rate: A practical and powerful approach to multiple testing},\
  }\href {https://doi.org/10.1111/j.2517-6161.1995.tb02031.x} {\bibfield
  {journal} {\bibinfo  {journal} {J. R. Stat. Soc. Ser. B}\ }\textbf {\bibinfo
  {volume} {57}},\ \bibinfo {pages} {289} (\bibinfo {year} {1995})}\BibitemShut
  {NoStop}%
\bibitem [{\citenamefont {Wang}\ and\ \citenamefont
  {Jacot}(2024)}]{wang2024implicit}%
  \BibitemOpen
  \bibfield  {author} {\bibinfo {author} {\bibfnamefont {Z.}~\bibnamefont
  {Wang}}\ and\ \bibinfo {author} {\bibfnamefont {A.}~\bibnamefont {Jacot}},\
  }\bibfield  {title} {\bibinfo {title} {Implicit bias of {SGD} in
  {$L_2$}-regularized linear {DNNs}: One-way jumps from high to low rank},\
  }in\ \href@noop {} {\emph {\bibinfo {booktitle} {International Conference on
  Learning Representations (ICLR)}}}\ (\bibinfo {year} {2024})\ \Eprint
  {https://arxiv.org/abs/2305.16038} {arXiv:2305.16038} \BibitemShut {NoStop}%
\bibitem [{\citenamefont {Kunin}\ \emph {et~al.}(2021)\citenamefont {Kunin},
  \citenamefont {Sagastuy-Brena}, \citenamefont {Ganguli}, \citenamefont
  {Yamins},\ and\ \citenamefont {Tanaka}}]{kunin2021neural}%
  \BibitemOpen
  \bibfield  {author} {\bibinfo {author} {\bibfnamefont {D.}~\bibnamefont
  {Kunin}}, \bibinfo {author} {\bibfnamefont {J.}~\bibnamefont
  {Sagastuy-Brena}}, \bibinfo {author} {\bibfnamefont {S.}~\bibnamefont
  {Ganguli}}, \bibinfo {author} {\bibfnamefont {D.~L.~K.}\ \bibnamefont
  {Yamins}},\ and\ \bibinfo {author} {\bibfnamefont {H.}~\bibnamefont
  {Tanaka}},\ }\bibfield  {title} {\bibinfo {title} {Neural mechanics: Symmetry
  and broken conservation laws in deep learning dynamics},\ }in\ \href@noop {}
  {\emph {\bibinfo {booktitle} {International Conference on Learning
  Representations (ICLR)}}}\ (\bibinfo {year} {2021})\ \Eprint
  {https://arxiv.org/abs/2012.04728} {arXiv:2012.04728} \BibitemShut {NoStop}%
\bibitem [{\citenamefont {Kobayashi}\ \emph {et~al.}(2024)\citenamefont
  {Kobayashi}, \citenamefont {Akram},\ and\ \citenamefont {von
  Oswald}}]{kobayashi2024weight}%
  \BibitemOpen
  \bibfield  {author} {\bibinfo {author} {\bibfnamefont {S.}~\bibnamefont
  {Kobayashi}}, \bibinfo {author} {\bibfnamefont {Y.}~\bibnamefont {Akram}},\
  and\ \bibinfo {author} {\bibfnamefont {J.}~\bibnamefont {von Oswald}},\
  }\bibfield  {title} {\bibinfo {title} {Weight decay induces low-rank
  attention layers},\ }in\ \href@noop {} {\emph {\bibinfo {booktitle} {Advances
  in Neural Information Processing Systems (NeurIPS)}}},\ Vol.~\bibinfo
  {volume} {37}\ (\bibinfo {year} {2024})\ pp.\ \bibinfo {pages} {4481--4510},\
  \Eprint {https://arxiv.org/abs/2410.23819} {arXiv:2410.23819} \BibitemShut
  {NoStop}%
\bibitem [{\citenamefont {Jacot}\ \emph {et~al.}(2025)\citenamefont {Jacot},
  \citenamefont {S{\'u}ken{\'i}k}, \citenamefont {Wang},\ and\ \citenamefont
  {Mondelli}}]{jacot2025wide}%
  \BibitemOpen
  \bibfield  {author} {\bibinfo {author} {\bibfnamefont {A.}~\bibnamefont
  {Jacot}}, \bibinfo {author} {\bibfnamefont {P.}~\bibnamefont
  {S{\'u}ken{\'i}k}}, \bibinfo {author} {\bibfnamefont {Z.}~\bibnamefont
  {Wang}},\ and\ \bibinfo {author} {\bibfnamefont {M.}~\bibnamefont
  {Mondelli}},\ }\bibfield  {title} {\bibinfo {title} {Wide neural networks
  trained with weight decay provably exhibit neural collapse},\ }in\ \href@noop
  {} {\emph {\bibinfo {booktitle} {International Conference on Learning
  Representations (ICLR)}}}\ (\bibinfo {year} {2025})\ \Eprint
  {https://arxiv.org/abs/2410.04887} {arXiv:2410.04887} \BibitemShut {NoStop}%
\bibitem [{\citenamefont {Ziyin}\ \emph
  {et~al.}(2025{\natexlab{a}})\citenamefont {Ziyin}, \citenamefont {Li},\ and\
  \citenamefont {Ueda}}]{ziyin2025noise}%
  \BibitemOpen
  \bibfield  {author} {\bibinfo {author} {\bibfnamefont {L.}~\bibnamefont
  {Ziyin}}, \bibinfo {author} {\bibfnamefont {H.}~\bibnamefont {Li}},\ and\
  \bibinfo {author} {\bibfnamefont {M.}~\bibnamefont {Ueda}},\ }\bibfield
  {title} {\bibinfo {title} {Noise balance and stationary distribution of
  stochastic gradient descent},\ }\href@noop {} {\bibfield  {journal} {\bibinfo
   {journal} {Phys. Rev. E}\ }\textbf {\bibinfo {volume} {111}},\ \bibinfo
  {pages} {065303} (\bibinfo {year} {2025}{\natexlab{a}})}\BibitemShut
  {NoStop}%
\bibitem [{\citenamefont {Ziyin}\ \emph
  {et~al.}(2025{\natexlab{b}})\citenamefont {Ziyin}, \citenamefont {Chuang},
  \citenamefont {Galanti},\ and\ \citenamefont {Poggio}}]{ziyin2025formation}%
  \BibitemOpen
  \bibfield  {author} {\bibinfo {author} {\bibfnamefont {L.}~\bibnamefont
  {Ziyin}}, \bibinfo {author} {\bibfnamefont {I.}~\bibnamefont {Chuang}},
  \bibinfo {author} {\bibfnamefont {T.}~\bibnamefont {Galanti}},\ and\ \bibinfo
  {author} {\bibfnamefont {T.}~\bibnamefont {Poggio}},\ }\bibfield  {title}
  {\bibinfo {title} {Formation of representations in neural networks},\ }in\
  \href@noop {} {\emph {\bibinfo {booktitle} {International Conference on
  Learning Representations (ICLR)}}}\ (\bibinfo {year} {2025})\ \Eprint
  {https://arxiv.org/abs/2410.03006} {arXiv:2410.03006} \BibitemShut {NoStop}%
\bibitem [{\citenamefont {Benaych-Georges}\ and\ \citenamefont
  {Nadakuditi}(2012)}]{benaychgeorges2012singular}%
  \BibitemOpen
  \bibfield  {author} {\bibinfo {author} {\bibfnamefont {F.}~\bibnamefont
  {Benaych-Georges}}\ and\ \bibinfo {author} {\bibfnamefont {R.~R.}\
  \bibnamefont {Nadakuditi}},\ }\bibfield  {title} {\bibinfo {title} {The
  singular values and vectors of low rank perturbations of large rectangular
  random matrices},\ }\href {https://doi.org/10.1016/j.jmva.2012.04.019}
  {\bibfield  {journal} {\bibinfo  {journal} {J. Multivar. Anal.}\ }\textbf
  {\bibinfo {volume} {111}},\ \bibinfo {pages} {120} (\bibinfo {year}
  {2012})}\BibitemShut {NoStop}%
\bibitem [{\citenamefont {Ding}(2020)}]{ding2020high}%
  \BibitemOpen
  \bibfield  {author} {\bibinfo {author} {\bibfnamefont {X.}~\bibnamefont
  {Ding}},\ }\bibfield  {title} {\bibinfo {title} {High dimensional deformed
  rectangular matrices with applications in matrix denoising},\ }\href
  {https://doi.org/10.3150/19-BEJ1129} {\bibfield  {journal} {\bibinfo
  {journal} {Bernoulli}\ }\textbf {\bibinfo {volume} {26}},\ \bibinfo {pages}
  {387} (\bibinfo {year} {2020})}\BibitemShut {NoStop}%
\bibitem [{\citenamefont {Pacco}\ and\ \citenamefont
  {Ros}(2023)}]{pacco2023overlaps}%
  \BibitemOpen
  \bibfield  {author} {\bibinfo {author} {\bibfnamefont {A.}~\bibnamefont
  {Pacco}}\ and\ \bibinfo {author} {\bibfnamefont {V.}~\bibnamefont {Ros}},\
  }\bibfield  {title} {\bibinfo {title} {Overlaps between eigenvectors of
  spiked, correlated random matrices: From matrix principal component analysis
  to random {G}aussian landscapes},\ }\href
  {https://doi.org/10.1103/PhysRevE.108.024145} {\bibfield  {journal} {\bibinfo
   {journal} {Phys. Rev. E}\ }\textbf {\bibinfo {volume} {108}},\ \bibinfo
  {pages} {024145} (\bibinfo {year} {2023})}\BibitemShut {NoStop}%
\bibitem [{\citenamefont {Baharav}\ \emph {et~al.}(2025)\citenamefont
  {Baharav}, \citenamefont {Nicol}, \citenamefont {Irizarry},\ and\
  \citenamefont {Ma}}]{baharav2025stacked}%
  \BibitemOpen
  \bibfield  {author} {\bibinfo {author} {\bibfnamefont {T.~Z.}\ \bibnamefont
  {Baharav}}, \bibinfo {author} {\bibfnamefont {P.~B.}\ \bibnamefont {Nicol}},
  \bibinfo {author} {\bibfnamefont {R.~A.}\ \bibnamefont {Irizarry}},\ and\
  \bibinfo {author} {\bibfnamefont {R.}~\bibnamefont {Ma}},\ }\href@noop {}
  {\bibinfo {title} {Stacked {SVD} or {SVD} stacked? {A} random matrix theory
  perspective on data integration}} (\bibinfo {year} {2025}),\ \Eprint
  {https://arxiv.org/abs/2507.22170} {arXiv:2507.22170} \BibitemShut {NoStop}%
\end{thebibliography}%

\end{document}